\documentclass[10pt,twocolumn,letterpaper]{article}

\usepackage[pagenumbers]{wacv} % To force page numbers, e.g. for an arXiv version

\definecolor{wacvblue}{rgb}{0.21,0.49,0.74}
\usepackage[pagebackref,breaklinks,colorlinks,allcolors=wacvblue]{hyperref}

\usepackage{amsmath}
\usepackage{amssymb}
\usepackage{mathtools}
\usepackage{amsthm}
\usepackage{dsfont}
\usepackage{makecell}
\usepackage{multirow}

\usepackage{algorithm}
\usepackage{algpseudocode}
\usepackage{xcolor}
\usepackage{color, colortbl}
\definecolor{LightGreen}{rgb}{0.56, 0.93, 0.56}
\newcommand{\CC}[1]{\cellcolor{LightGreen}}

\usepackage[english]{babel}

\usepackage[toc,header]{appendix}
\usepackage{titletoc}

\newcommand{\shasam}{\textsc{SHaSaM}}

\def\wacvPaperID{1704} % *** Enter the WACV Paper ID here
\def\confName{WACV}
\def\confYear{2026}

\title{\shasam: Submodular Hard Sample Mining for Fair Facial Attribute Recognition}

\author{Anay Majee\\
The University of Texas at Dallas\\
{\tt\small anay.majee@utdallas.edu}
\and
Rishabh Iyer\\
The University of Texas at Dallas\\
{\tt\small rishabh.iyer@utdallas.edu}
}

\begin{document}
\maketitle

\begin{abstract}
Deep neural networks often inherit social and demographic biases from annotated data during model training, leading to unfair predictions, especially in the presence of sensitive attributes like race, age, gender etc.
Existing methods fall prey to the inherent data imbalance between attribute groups and inadvertently emphasize on sensitive attributes, worsening unfairness and performance.  
To surmount these challenges, we propose \textbf{\shasam} (\textbf{S}ubmodular \textbf{Ha}rd \textbf{Sa}mple \textbf{M}ining), a novel combinatorial approach that models fairness-driven representation learning as a submodular hard-sample mining problem. 
Our two-stage approach comprises of \shasam-MINE, which introduces a submodular subset selection strategy to mine hard positives and negatives — effectively mitigating data imbalance, and \shasam-LEARN, which introduces a family of combinatorial loss functions based on Submodular Conditional Mutual Information to maximize the decision boundary between target classes while minimizing the influence of sensitive attributes. 
This unified formulation restricts the model from learning features tied to sensitive attributes, significantly enhancing fairness without sacrificing performance. 
Experiments on CelebA and UTKFace demonstrate that \shasam\ achieves state-of-the-art results, with up to 2.7 points improvement in model fairness (Equalized Odds) and a 3.5\% gain in Accuracy, within fewer epochs as compared to existing methods.\looseness-1
\end{abstract}

% START of paper
\section{Introduction}
\begin{figure}[h]
        \centering
        \includegraphics[width=0.85\columnwidth]{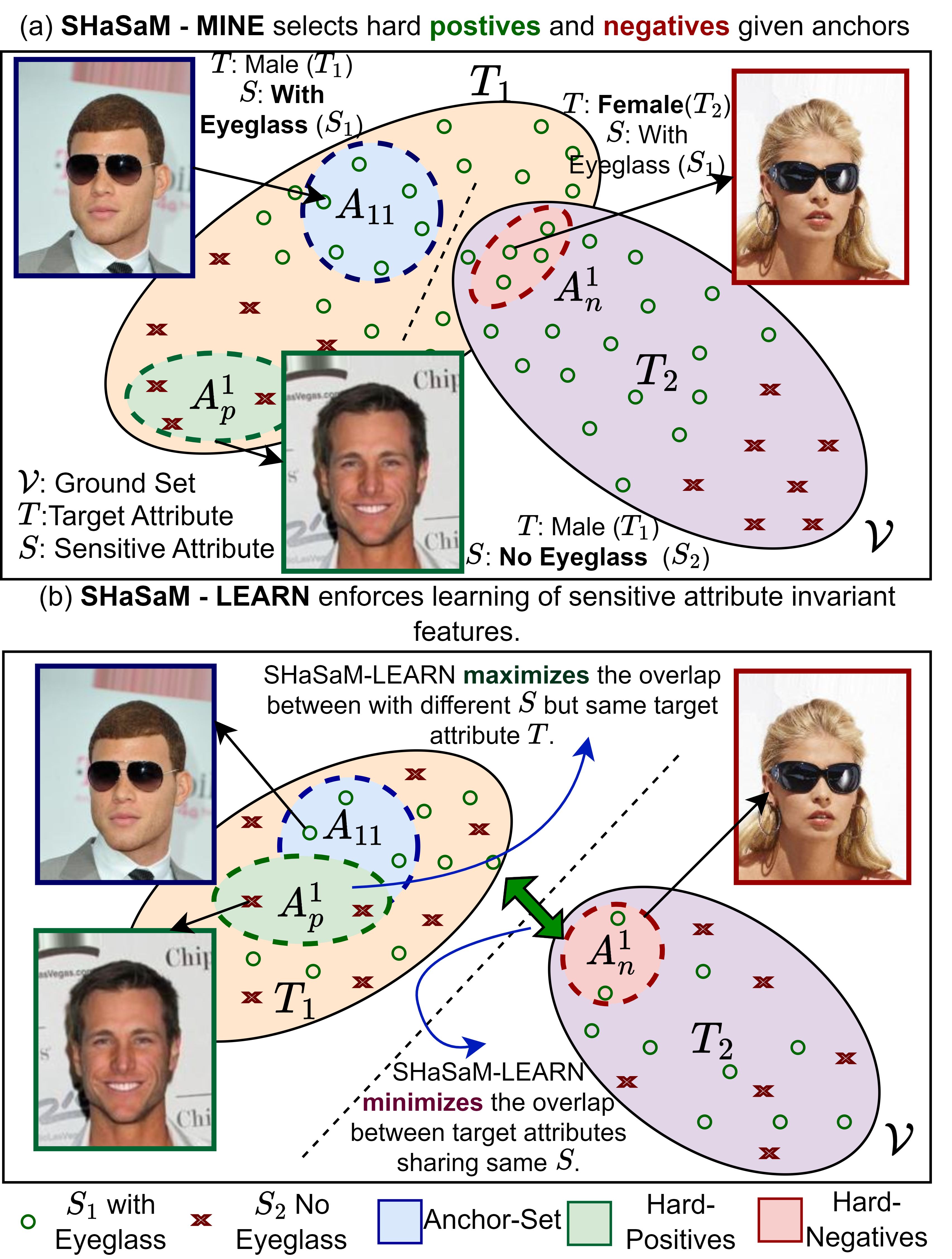}
        \caption{\textbf{Illustration of \shasam} - Given a target $T$ (gender) and sensitive attribute $S$ (eyeglasses) set, (a) \shasam-MINE selects hard-positives $A_p^1$ and hard-negatives $A_n^1$ given anchor set $A_{11}$. (b) \shasam-LEARN enforces a decision boundary between target attributes $T$ invariant to the sensitive attribute $S$.\looseness-1}
        \label{fig:intro_shasam}
        \vspace{-4ex}
\end{figure}

Training modern Deep Neural Networks (DNNs)~\cite{resnet, vit} on curated datasets often incorporates the social and demographic biases of human annotators, which can result in unfair predictions during inference. While removing these biases is challenging as they reflect real-world scenarios, various approaches aim to mitigate their impact by encouraging DNNs to learn fair feature representations. Contrastive learning~\cite{simclr, supcon, moco} methods, for example, have been applied to group instances with similar target attributes and separate those with different attributes to enhance fairness. However, these approaches could induce biased responses~\cite{fscl, Shen2021ContrastiveLF} in the presence of sensitive attributes like race, age, gender etc. \citet{fscl} attributes this to the inadvertent learning of irrelevant features linked to sensitive attributes, weakening the decision boundary between target attributes and impacting fairness in downstream tasks. Additionally, real-world datasets~\cite{celeba, utkface, karkkainen2021fairface} often exhibit significant imbalances in target and sensitive attributes due to the curation process, introducing biases that favor more abundant attributes and exacerbate unfair predictions. While imbalance among attribute groups has been addressed in tasks like image recognition~\cite{glmc, gpaco, score}, imbalances in sensitive attributes have received limited attention, despite introducing high intra-group variance~\cite{fscl} within each target class and ultimately degrading downstream performance.\looseness-1

To address challenges in fair facial attribute recognition, we propose \textbf{\shasam}, a combinatorial (set-based) method that frames fairness-driven representation learning as a \textbf{S}ubmodular \textbf{Ha}rd \textbf{Sa}mple \textbf{M}ining task.
Motivated by the recent success of \cite{score, smile} in introducing combinatorial objectives for representation learning, \textit{we introduce a set-based (combinatorial) viewpoint in facial attribute recognition} (see \cref{sec:prob_def}) allowing us to leverage the benefits of submodular functions~\cite{fujishige} to learn target attribute representations invariant to the sensitive ones.
At first, \shasam-MINE selects hard-positives and hard-negatives for each diverse set of anchor examples (all with the same target and sensitive attribute) as shown in \cref{fig:intro_shasam}(a), modeling hard-sample mining as a combinatorial selection problem~\cite{submod_diversity, vid_sum_2019, talisman}. Following the intuition in \citet{fscl}, \shasam-MINE mines hard-positives with the same target but different sensitive attributes and hard-negatives with the same sensitive but different target attributes, thus \textbf{balancing positives and negatives per training iteration to mitigate attribute group imbalances}.\looseness-1

Finally, \shasam-LEARN introduces a family of objectives based on Submodular Conditional Mutual Information (CMI)~\cite{prism}. Minimizing CMI in \shasam-LEARN leverages the diversity-maximizing~\cite{submod_diversity} and cooperation-enhancing~\cite{submod_cooperation} properties of submodular functions in a unified objective. As shown in \cref{fig:intro_shasam}(b), we simultaneously maximize the information overlap between anchors and hard-positives and minimizes inter-group separation between anchors and hard-negatives mined by \shasam-MINE. This unified framework \textbf{restricts the model from learning features linked to sensitive attributes, significantly enhancing fairness} without sacrificing downstream performance. In practice, \shasam-MINE and \shasam-LEARN are sequentially invoked at each training iteration on the facial attribute recognition benchmarks~\cite{fscl, tian2024fairvit} demonstrating the following major contributions:\looseness-1

\begin{itemize}[leftmargin=*]
\item \textbf{\shasam\ introduces a novel set-based combinatorial viewpoint in fair facial attribute classification} by modeling fair attribute recognition as a \textbf{submodular hard-sample mining problem} using a two-stage framework encouraging the learning of fair feature embeddings corresponding to target attributes without biasing on sensitive ones during model training.\looseness-1
\item Components of \shasam, namely \shasam-MINE \textbf{employs a novel submodular subset selection strategy} to mine hard-positives and hard-negatives mitigating the impedence of \textit{data imbalance among attribute groups} (detailed in \cref{sec:shasam_mine}).\looseness-1
\item \textbf{\shasam-LEARN introduces a family of combinatorial objectives based on  Submodular Conditional Mutual Information which presents an unified approach to improve \textit{fairness}} in model predictions by minimizing the information overlap between target classes sharing the same sensitive attribute while maximizing the overlap within target classes sharing different sensitive attributes.\looseness-1
\item Overall, \shasam\ demonstrates up to \textbf{2.7 points improvement in fairness metrics (quantified through Equalized Odds (EO))}, achieving up to 3.5\% boost in Top-1 Acc. on popular facial attribute recognition benchmarks like CelebA and UTKFace within fewer training epochs.\looseness-1
\end{itemize}

\section{Related Work}
\label{sec:rel_work}
Ensuring fairness in machine learning, particularly in facial attribute classification, has become pivotal due to the significant progress has been made in mitigating biases while maintaining high model performance.
Early efforts focused on adversarial learning and dataset adjustments. \citet{wadsworth2018advererial} pioneered adversarial learning to reduce bias in decision-making tasks, while \citet{raff2018reversal} proposed gradient reversal techniques to mitigate discrimination in neural networks; both demonstrated the effectiveness of adversarial approaches during training. To address gender bias in facial recognition, \citet{zhao2017corpus} proposed a debiasing technique that generates gender-neutral face embeddings, significantly improving fairness. Similarly, \citet{kim2019notlearn} introduced an adversarial framework to prevent models from learning biased data representations. Disentanglement methods were also explored, with \citet{creager2019disentanglement} separating sensitive and non-sensitive features in learned representations, and \citet{arjovsky2019invariant} introducing Invariant Risk Minimization (IRM) to learn representations invariant to data distribution shifts. 
Dataset balancing methods have been adopted to mitigate attribute imbalance and reduce inter-attribute bias~\cite{karkkainen2021fairface, raff2018reversal, romano2020resampling}. FairFace~\cite{karkkainen2021fairface} introduced a dataset designed to address imbalances in race, gender, and age within facial attribute classification tasks. Techniques like gradient reversal~\cite{raff2018reversal} and resampling of rare attributes~\cite{romano2020resampling} achieve lower equalized odds, ensuring equitable outcomes across diverse demographic groups.\looseness-1

Generative models have also played a prominent role in promoting fairness. \citet{shen2020interfacegan} introduced InterfaceGAN, leveraging Generative Adversarial Networks (GANs) to manipulate latent space representations of facial attributes such as age and gender, allowing fine-tuned control over fairness in attribute manipulation tasks. Orthogonal disentanglement between task-specific and sensitive attributes, as shown by \citet{Sarhan2020disentanglement}, improved fairness across downstream classification tasks. Additionally, \citet{wang2020bias} explored various strategies for bias mitigation in visual recognition, including resampling and re-weighting techniques, contributing significantly toward mitigating bias in large-scale vision systems.

Recent advancements in fairness-aware representation learning have focused heavily on contrastive learning~\cite{simclr,supcon,moco,mocov2}. \citet{Park2021disentanglement} proposed a fairness-aware information alignment method to disentangle sensitive attributes, enhancing fairness in facial attribute classification. \citet{ramaswamy2021latentdebiasing} applied latent space de-biasing techniques to eliminate correlations between target outputs and sensitive features without sacrificing accuracy. \citet{jung2021distillation} introduced Fair Feature Distillation, a contrastive learning method aimed at learning invariant representations, further enhancing fairness in visual recognition. Latent space smoothing, proposed by \citet{peychev2022smoothing}, ensures that small input variations do not lead to disproportionately large changes in model predictions. This approach, along with fairness-preserving latent representations in \citet{roh2023drfairness}, balances fairness and model performance. Transformer-based methods have also been explored; \citet{qiang2023fairness} and \citet{Sudhakar2023MitigatingBI} introduced Debiased Self-Attention and TADeT, respectively, aiming to eliminate biases in Vision Transformers. More fundamentally, \citet{tian2024fairvit} modified the attention head architecture in Vision Transformers~\cite{dosovitskiy2021vit} to induce fairness. Although transformer-based approaches improve both fairness and performance, they come with significant computational overhead.

Representation learning techniques have thus shown remarkable promise in addressing the fairness challenges inherent in facial attribute classification. From early disentanglement methods~\cite{creager2019disentanglement} to recent advancements in contrastive learning~\cite{Shen2021ContrastiveLF, fscl, jung2021distillation} and latent space smoothing~\cite{peychev2022smoothing}, these techniques have transformed how fairness is incorporated into facial recognition models. Fair representation learning has evolved into a comprehensive field where sensitive attributes are decoupled from task-specific features, ensuring unbiased predictions while retaining strong performance. As demonstrated by \citet{madras18advfair}, adversarial learning can be effectively applied to achieve fair and transferable representations, laying the foundation for future work in this critical area. \citet{Shen2021ContrastiveLF} further advanced this field by incorporating contrastive learning approaches that significantly reduce bias in large-scale visual systems, solidifying the role of representation learning as a cornerstone in building equitable Artificial Intelligence (AI) systems.\looseness-1

\section{Method}
\label{sec:method}
\subsection{Problem Definition}
\label{sec:prob_def}
We consider a dataset $D \in D^{train} \cup D^{test}$ with a training set $D^{train}$ consisting of $N$ labeled samples such that $\{<x_i, t_i, s_i>\}_{i = 1}^{N}$. 
Here, $x_i$ represents the raw data while $t_i$ and $s_i$ jointly compose the label indicating the target and sensitive attribute for $x_i$ in $D^{train}$.  
Introducing a combinatorial viewpoint as in \cite{score, smile} we formulate $D_{train}$ as a collection of sets (each containing a set of datapoints) based on the target attribute $T_i$ or the sensitive attribute $S_i$ such that $D_{train} = \mathcal{V} = \cup_{i =1}^N T_i = \cup_{j = 1}^K S_j$, where $N \neq K$. 
A feature extractor $F(A; \theta)$, maps all instances in the input set $A$ into a lower dimensional feature space and a classifier $Clf(F(A); \theta)$ maps the learnt feature representations to its corresponding target attribute $i \in N$ irrespective of the sensitive attribute $j \in K$.
The optimization task~\cite{fscl, Shen2021ContrastiveLF} is to maximize the probability of all elements in $A$ belonging to the target class $T_i$ without biasing on the sensitive attribute $S_j$.\looseness-1

\begin{figure}
        \centering
        \includegraphics[width=0.9\columnwidth]{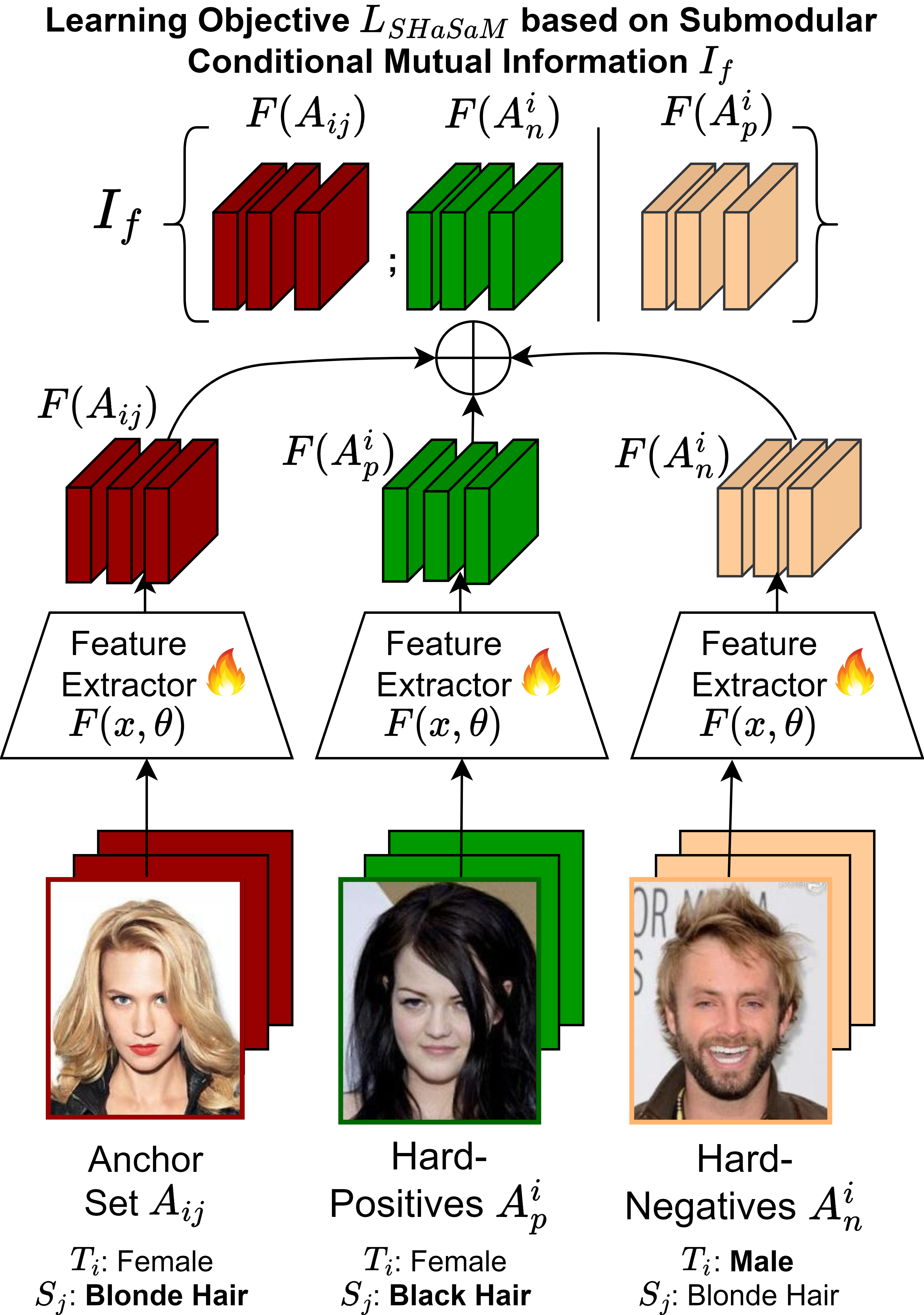}
        \caption{\textbf{Training Strategy in \shasam-LEARN} which learns parameters of $F(x, \theta)$ by minimizing a novel combinatorial objective $L_{\shasam}$ to learn features invariant to sensitive attributes.}
        \label{fig:main_fig}
        \vspace{-4ex}
\end{figure}

\subsection{Preliminaries: Submodularity}
\label{sec:submod_prelims}
Submodular functions~\cite{fujishige,iyer2015polyhedral,bilmes2022submodularity} are set functions characterized by a natural diminishing returns property. Specifically, a set function \( f: 2^\mathcal{V} \rightarrow \mathbb{R} \), defined over a ground set \(\mathcal{V}\), is submodular if it satisfies the inequality \( f(A_i) + f(A_j) \geq f(A_i \cup A_j) + f(A_i \cap A_j) \) for all subsets \( A_i, A_j \subseteq \mathcal{V} \)~\citep{fujishige}. 
These functions have been extensively researched in areas such as data subset selection~\citep{killamsetty_automata, prism, jain2023efficient,durga2021training,killamsetty2022nested}, active learning~\citep{wei15_subset, talisman, Beck2021EffectiveEO, Kaushal_2019}, and video summarization~\citep{vid_sum_2019, kaushal2019framework, kaushal2021vid_summary}. 
Traditionally, subset selection and summarization tasks are modeled as a discrete optimization problem through submodular maximization~\cite{fujishige, Nemhauser1978} under a knapsack constraint~\cite{Nemhauser1978}. This can be fairly approximated with a $(1 - e^{-1})$ constant factor guarantee~\cite{Nemhauser1978} using greedy optimization techniques~\cite{Nemhauser1978,Mirzasoleiman2015lazierthanlazy}.
Maximizing \textbf{Submodular Information Functions} (SIMs)~\cite{iyer2021submodular,iyer2021generalized,bilmes2022submodularity} $f(A)$ like Facility-Location, Graph-Cut etc. promotes selection of diverse examples within a set $A$, while maximizing \textbf{Submodular Mutual Information Functions} (SMIs) $I_f(A;Q)$ selects examples that share maximum information in $A \cap Q$. Complimentary to SMI, \textbf{Submodular Conditional Gain} (SCG) $H_f(A | Q)$ selects examples in $A$ with maximum dissimilarity to $Q$, while \textbf{Submodular Conditional Mutual Information} (SCMI) $I_f(A; Q | P)$ selects examples in $A$ similar to $Q$ but dissimilar to $P$. Here, $P$ and $Q$ are disjoint sets. We include more details in the Appendix.
Interestingly, recent approaches \cite{score, smile} have also applied combinatorial functions as learning objectives in continuous optimization problems. Specifically, \citet{score} introduces combinatorially inspired loss functions (in this case SIMs) which model intra-group compactness (cooperation) when minimized and inter-group separation when maximized. Additionally, \cite{smile} introduces SMI as loss function to model interactions between abundant and rare examples in data-efficient (few-shot) representation learning tasks. 
In our paper, we exploit a mixture of discrete and continuous optimization tasks through a novel combinatorial approach to induce fairness in representation learning.\looseness-1

\subsection{\shasam : Submodular Hard Sample Mining}
\label{meth:shasam}
Existing works in fair attribute recognition (refer \cref{sec:rel_work}) highlight two major challenges in the domain - (1) Overcoming large \textit{imbalance} not only between target attributes but also between sensitive attributes (with the same target attribute), and (2) Learning discriminative \textit{feature representations} for each target attribute $T_i \in \mathcal{V}$ without biasing on the sensitive attribute $S_j \in \mathcal{V}$.
In a quest to learn a target attribute $T_i$, examples with rare sensitive attributes $S_j$ belonging to $T_i$ act as hard positives while examples belonging to $\mathcal{V} \setminus T_i$ (with the same $S_j$) act as hard negatives~\cite{lsl}.

In this paper, we introduce a novel unified approach \shasam\ by modeling fairness driven representation learning as a combinatorial hard sample mining problem. 
We achieve this through two key steps - (1) \textbf{\shasam-MINE} (detailed in \cref{sec:shasam_mine}) which mines hard positives $A_p^i$ and hard negatives $A_n^i$ given a anchor set $A_{ij}$ mined from the dataset $\mathcal{V}$, and (2) \textbf{\shasam-LEARN} (detailed in \cref{sec:shasam_learn}) which introduces a family of novel combinatorial learning objectives based on Submodular Conditional Mutual Information to jointly maximizes cluster overlap between $A_{ij}$ and $A_p^i$ (which share different $S_j$'s) while maximizing the decision boundary between $A_{ij}$ and $A_n^i$ (which share the same $S_j$ but belong to different $T_i$'s).

We define the selection of a candidate set $A$ from a collection $Q$ as a $\texttt{softmax}$ over a selection function $\mathcal{F}$ as shown in \cref{eq:softmax}. This definition of $\texttt{softmax}$ serves as an approximation of the indicator over the $\texttt{argmax}$ function. This is a smooth approximation modeling the selection of the most relevant example (positive / negative) from $Q$ depending on the choice of $\mathcal{F}$ similar to the definition in \citet{lsl}.\looseness-1
\begin{align}
    \texttt{softmax}(\mathcal{F}(.), A, Q) = \frac{e^{\mathcal{F}(A)}}{\sum_{p \in Q} e^{\mathcal{F}(p)}} \approx \mathds{1}_{\underset{A \subseteq Q}{\texttt{argmax}} \mathcal{F}(A)}
    \label{eq:softmax}
\end{align}
Now, given a diverse set of anchors $A_{ij} \in  \{T_i \cap S_j\},  \forall i,j \in |\mathcal{V}|, T_i \cap S_j \neq \phi$, we mine hard-positives $A_p^i$ from $T_i \cap \overline{S_j}$ which maximizes the Submodular Conditional Gain (SCG) $H_f(.| A_{ij})$ denoted as $\texttt{softmax}(H_f(.|A_{ij}), A_p^i, T_i \cap \overline{S_j}))$. 
Similarly, we mine hard-negatives $A_n^i$ from $\mathcal{V} \setminus T_i \cap S_j$ which maximizes the Submodular Mutual Information (SMI) $I_f(. ; A_{ij})$ denoted as $\texttt{softmax}(I_f(.;A_{ij}), A_n^i, \mathcal{V} \setminus T_i \cap S_j))$. Here, $I_f$ and $H_f$ act as selection functions.
The mined anchors, hard-positives and hard-negatives mined at each training iteration train the parameters of the feature extractor $F(.; \theta)$ by minimizing a novel combinatorial learning objective $L_{\shasam}(A_{ij}, A_p^i, A_n^i; \theta)$ defined in \shasam-LEARN (refer \cref{sec:shasam_learn}). Jointly, we can thus summarize the hard-sample sample mining task as a joint selection and representation learning problem described in \cref{eq:formulation}.\looseness-1
\begin{align}
\begin{split}
   L_{\shasam}(\theta) = &\sum_{\forall i,j \in |\mathcal{V}|} 
   \underbrace{\texttt{softmax}(H_f(.|A_{ij}), A_p^i, T_i \cap \overline{S_j}))}_\text{(A) Selecting Hard Positives in $T_i \cap \overline{S_i}$} \\
     &\times \underbrace{\texttt{softmax}(I_f(.;A_{ij}), A_n^i, \mathcal{V} \setminus T_i \cap S_j))}_\text{(B) Selecting Hard Negatives in $\mathcal{V} \setminus T_i \cap S_j$} \\
   & \times \underbrace{L_{\shasam}(A_{ij}, A_p^i, A_n^i ; \theta)}_\text{(C) \shasam-LEARN} 
\end{split}
   \label{eq:formulation}
\end{align}
We minimize $L_{\shasam}(\theta)$ during model training. Note that $L_{\shasam}(\theta) \approx L_{\shasam}(\hat{A_{ij}}, \hat{A_p^i}, {A_n^i} ; \theta)$ where $A_p^i = \texttt{argmax}_{A \subseteq T_i \cap \overline{S_j}} H_f(A | A_{ij})$ and $A_n^i = \texttt{argmax}_{A \subseteq  \mathcal{V} \setminus T_i \cap S_j} I_f(A ; A_{ij})$. Similarly, $\nabla L_{\shasam}(\theta) \approx \nabla L_{\shasam}(\hat{A_{ij}}, \hat{A_p^i}, {A_n^i} ; \theta)$ since the gradients of the other terms involving the \texttt{softmax} is close to $0$. This reduces the learning problem to minimizing $L_{\shasam}$ on the mined anchors, hard positives and negatives as shown in \cref{fig:main_fig}.
The aforementioned phenomenon further enables us to model the selection functions in (A) and (B) as submodular maximization~\cite{Nemhauser1978, fujishige} tasks, $\texttt{argmax}_{A \subseteq \mathcal{Q}, |A| \leq k} \mathcal{F}(A)$, under a cardinality constraint $|A| \leq k$ (selecting $k$ examples in $A$), given $\mathcal{F}$ is inherently submodular. This is further described in \cref{sec:shasam_mine}.\looseness-1

\subsection{\shasam-MINE}
\label{sec:shasam_mine}
Given the ground set $\mathcal{V}$ (as defined in \cref{sec:prob_def}) and a budget $\mathtt{k}$ we select anchors, hard positives and negatives through a novel selection strategy \shasam-MINE as elucidated in \cref{alg:dss}. 
For every iteration $t$ in an epoch we choose candidate target and sensitive attributes $i$ and $j$ (lines 16, 17 in \cref{alg:dss}), $\forall i,j \in |\mathcal{V}|$ such that samples in $T_i \cap S_j \neq \phi$.
Next, we mine anchor set $A_{ij}$ for the selected ($T_i$, $S_j$) pair through submodular maximization of the Log-Determinant (LogDet) function $f(A_{ij})$ over each subset $A_{ij} \in \mathcal{V}$ as shown in line 19 of \cref{alg:dss}, under a cardinality constraint. 
% \begin{align}
%    A_{ij} \leftarrow \underset{\substack{A_{ij} \in T_i \cap S_j \\ |A_{ij}| \leq k}}{\max} f(A_{ij}), \forall i,j \in |\mathcal{V}|
%    \label{eq:query_selection}
% \end{align}
\cref{fig:shasam_mine_res}(a) shows that maximizing LogDet selects diverse examples~\cite{vid_sum_2019} from $T_i \cap S_j$ in $A_{ij}$. Following this step, we perform a two stage selection on $\mathcal{V}$ to identify hard-positives $A_p^i$ and hard-negatives $A_n^i$ relative to the choice of a diverse anchor set $A_{ij}$.\looseness-1

\noindent \textbf{Hard-Positive samples} $A_p^i$ are selected by \textit{maximizing the SCG function $H_f(A_p^i | A_{ij})$ over all samples in $T_i \cap \overline{S_j}$} to mine exactly $\mathtt{k}$ (represented as \textit{budget}) as shown in line 20 of \cref{alg:dss}. 
Maximizing SCG results in selection of samples which have maximum dissimilarity to the ones in $A_{ij}$ \cite{prism} but share the same $T_i$. Following \citet{prism}, we adopt the Facility Location (FL) as the underlying function to compute the SCG resulting in selection of representative \textit{hard-positives} in $A_p^i$ (share the same class label but different sensitive attribute with $A_{ij}$) located at the cluster boundary of $T_i$ as shown in \cref{fig:shasam_mine_res}(b).

\begin{algorithm}[tb]
\caption{Representation Learning stage in \shasam} \label{alg:dss}
\begin{algorithmic}[1]
\Require Dataset $D^{train}$, Model $F$, Initialized model parameters $\theta$, No. of epochs $E$, Batch size $b$, budget $\mathtt{k}$

\State \textcolor{blue}{\textit{/** Training Loop in \shasam **/}} 
\State $\mathcal{V}_{prev} = \phi$
\For{$e = 1, 2, \dots, E$}
    \State $\mathcal{V} \leftarrow \texttt{RANDOM-SAMPLE}(D^{train}, 0.2) \setminus \mathcal{V}_{prev}$ 
    \For{$t = 1, 2, \dots, |\mathcal{V}|/b$}
        \State $\{A, P, N\} = \texttt{\shasam-MINE}(\mathcal{V}, \mathtt{k})$
        \State \textcolor{gray}{\textit{/* Forward pass */}}
        \State $h_A, h_P, h_N \leftarrow F(A; \theta), F(P; \theta), F(N; \theta)$
        \State $L(\theta), \nabla_{\theta} \leftarrow \texttt{\shasam-LEARN}(h_A, h_P, h_N, \theta)$
        \State $\theta := \theta - \eta.\nabla_{\theta}$ \Comment{\textcolor{blue}{\textit{// Backward Pass}}}
    \EndFor
    \State $\mathcal{V}_{prev} = \mathcal{V}$
\EndFor

\State \textcolor{blue}{\textit{/**Hard Sample Mining in \shasam-MINE**/}} 
\Function{\shasam-MINE}{$\mathcal{V}, \mathtt{k}$}
    \State $i = \texttt{RANDOM-CHOICE}([1, |\mathcal{V}|])$  
    \State $j = \texttt{RANDOM-CHOICE}([1, |\mathcal{V}|])$ 
    \State $\textbf{Assert } T_i \cap S_j \neq \phi$
    \State $A_{ij} \leftarrow \underset{\substack{A_{ij} \in T_i \cap S_j \text{ , } |A_{ij}| \leq \mathtt{k}}}{\texttt{argmax}} f(A_{ij})$
    \State $A_p^i \leftarrow \underset{\substack{A_p^i \subseteq T_i \cap \overline{S_j} \text{ , } |A_p^i| \leq \mathtt{k}}}{\texttt{argmax}}  H_f(A_p^i | A_{ij})$
    \State $A_n^i \leftarrow \underset{\substack{A_n^i \subseteq \mathcal{V} \setminus T_i \cap S_j \text{ , } |A_n^i| \leq \mathtt{k}}}{\texttt{argmax}} I_f(A_n^i ; A_{ij})$
    \State \Return $\{A_{ij}, A_p^i, A_n^i\}$
\EndFunction

\State \textcolor{blue}{\textit{/**Learning Objective in\shasam-LEARN**/}} 
\Function{\shasam-LEARN}{$\hat{A_{ij}}, \hat{A_p^i}, \hat{A_n^i}, \theta$}
    \State $L_{\shasam}(\theta) = \frac{1}{N_f(\hat{A_{ij}})} I_f\left(\hat{A_{ij}}; \hat{A_n^i} | \hat{A_p^i}; \theta\right)$
    \State $\nabla L_{\shasam}(\theta) = \nabla L_{\shasam}(\hat{A_{ij}}, \hat{A_p^i}, \hat{A_n^i} ; \theta)$
    \State \Return $L_{\shasam}(\theta)$, $\nabla L_{\shasam}(\theta)$
\EndFunction
\end{algorithmic}
\end{algorithm}

\noindent \textbf{Hard-Negative samples} in $A_n^i \subseteq \mathcal{V} \setminus T_i \cap S_j$ are selected such that it \textit{maximizes the SMI function $I_{f}(A_n^i ; A_{ij})$ between the anchor set $A_{ij}$ and $A_n^i$} as shown in line 21 of \cref{alg:dss} under the same cardinality constraint. 
Adopting Facility-Location as the underlying submodular function, the resultant subset $A_n^i$ upon maximizing SMI contains examples most similar to $A_{ij}$~\cite{prism, orient}, sharing the same sensitive attribute label $S_j$ but with a different target attribute from the samples in $A_{ij}$. 

Note that all maximization steps follow discrete optimization through the Lazy-Greedy algorithm~\cite{Mirzasoleiman2015lazierthanlazy} (detailed in Appendix) under a cardinality constraint (same across tasks).  
Interestingly, application of the cardinality constraint results in balanced selection $|A_{ij}| = |A_p^i| = |A_n^i| = \mathtt{k}$ in each input batch, \textit{significantly overcoming the effect of class-imbalance between attribute groups}.\looseness-1

\subsection{\shasam-LEARN}
\label{sec:shasam_learn}
Given a ground set $\mathcal{V}$ (as defined in \cref{sec:prob_def}), a submodular function $f(A, \theta)$ over a set $A$ and an input batch of examples from \shasam-MINE $\{A_{ij}, A_p^i, A_n^i\}$ at each iteration, we define a loss function $L_{\shasam}(\theta)$ in \shasam-LEARN (refer \cref{alg:dss}) as an instance of Submodular Conditional Mutual Information (SCMI) function $I_f(A_{ij}; A_n^i | A_p^i, \theta) = f(A_{ij} \cup A_p^i) + f(A_n^i \cup A_p^i) - f(A_{ij} \cup A_p^i \cup A_n^i) - f(A_p^i)$ which jointly minimizes the feature overlap between $A_{ij}$ and $A_n^i$ while maximizing the mutual information between $A_{ij}$ and the hard-positives $A_p^i$ as shown in \cref{eq:shasam_cmi}. Here, $N_f(A_{ij})$ is the normalization constant.
\begin{align}
   L_{\shasam}(\theta) &= \sum_{\forall i,j \in |\mathcal{V}|} \frac{1}{N_f(A_{ij})} I_f\left(A_{ij}; A_n^i | A_p^i; \theta\right)
   \label{eq:shasam_cmi}
\end{align}
This formulation largely differs from \cite{score, smile} which introduce combinatorial objectives to model either intra-group compactness through SIMs~\cite{score} or inter-group separation through SMI~\cite{smile} without unifying these properties into a single formulation as shown in \shasam.
Further, \cref{sec:shasam_mine} points out that $A_p^i$ and $A_n^i$ are located at the cluster boundary of disjoint feature clusters. Thus, minimizing $L_{\shasam}$ \textit{increases the decision boundary} between them. Concurrently, minimizing $L_{\shasam}$ enforces an increased feature overlap between $A_{ij}$ and $A_p^i$ which belong to the same $T_i$ but differ in the sensitive attribute label $S_j$. This results in a \textit{compact feature cluster for $T_i$ invariant to the sensitive attribute enforcing fairness} in the learning process.

We model interactions between samples within/across sets using the cosine similarity kernel $S_{ij}(\theta) = \frac{F(x_{i}, \theta)^{\text{T}} \cdot F(x_{j}, \theta)}{||F(x_{i}, \theta)|| \cdot ||F(x_{j}, \theta)||}$, $\forall i,j \in |\mathcal{V}|$ and aggregate these interactions to compute the SCMI objective. By varying the submodular function $f(A, \theta)$ between Facility-Location (FL) and LogDet we introduce a family of objectives - \shasam-FLCMI and \shasam-LogDetCMI summarized in \cref{tab:instances_shasam_learn} with detailed derivations in the Appendix. 
Empirical evidence in \cref{sec:experiments} suggest that \shasam-FLCMI based objective outperforms other instances alongside SoTA approaches in fair attribute classification due to its inherent self-balancing~\cite{score} property alongside being insensitive to the presence of sensitive attributes during training.\looseness-1

\begin{table}[t]
\caption{\textbf{Instances of learning objectives in \shasam-LEARN} obtained by varying the submodular function $f(A)$ over a set $A$. Here $S$ indicates the cosine similarity kernel.}
\centering
\small
\resizebox{1.0\columnwidth}{!}{
\begin{tabular}{c|c}
\hline 
Instance Name & $L_{\shasam}(A_{ij}, A_p^i, A_n^i, \theta)$ \\
\hline 
\shasam- & \multirow{ 2}{*}{\makecell[c]{
$\displaystyle \sum_{i,j \in |\mathcal{V}|} \frac{1}{3|A_{ij}|} \left( \max \left( \min \left( \max_{a \in A_{ij}} S_{ia}, \max_{n \in A_n^i} S_{in} \right) - \max_{p \in A_p^i} S_{ip}, 0 \right) \right)$
} }\\
FLCMI & \\
& \\
& \multirow{ 2}{*}{\makecell[c]{
$\displaystyle \sum_{i,j \in |\mathcal{V}|} \frac{1}{3|A_{ij}|} \log \frac{\det \left( I - S_{A_n^i}^{-1} S_{{A_n^i},{A_p^i}} S_{A_p^i}^{-1} S_{{A_n^i},{A_p^i}}^T \right)}{\det \left( I - S_{A_{ij} \cup {A_p^i}}^{-1} S_{A_{ij} \cup {A_p^i},{A_n^i}} S_{A_n^i}^{-1} S_{A_{ij} \cup {A_p^i},{A_n^i}}^T \right)}$
} } \\
\shasam- & \\
LogDetCMI & \\
& \\
\hline
\end{tabular}}
\label{tab:instances_shasam_learn}
\end{table}

\section{Experiments}
\label{sec:experiments}
\paragraph{Datasets} We perform our experiments on two benchmark datasets adopted from \cite{fscl, tian2024fairvit, Shen2021ContrastiveLF} - 

\noindent (1) \textbf{CelebA}~\cite{celeba} contains approximately 200k facial images annotated with 40 binary attributes. We follow \cite{fscl} and designate 'male' (m) and 'young' (y) as the sensitive attributes and select target attributes that exhibit the highest Pearson correlation with these sensitive attributes. To ensure reliable evaluation, we manually exclude attributes that are extremely correlated; for instance, the 'heavy-makeup' attribute. Consequently, we utilize three single target attributes: 'attractiveness' (a), 'big nose' (b), and 'bags under eyes' (e), as well as a pair of target attributes - {'big nose', 'bags under eyes'}.\looseness-1

\begin{table*}[ht]
\caption{\textbf{Classification results on CelebA} measured in terms of Top-1 Accuracy (Acc.) and equalized odds (EO) by varying the target $T$ and sensitive attributes $S$. Here, $a$, $b$, $e$, $m$, and $y$ denote attractiveness, big nose, bags-under-eyes, male, and young, respectively. All results are averaged over three independent runs. * indicates our re-implementations.}
\centering
\resizebox{\textwidth}{!}{
\begin{tabular}{l|cc|cc|cc|cc|cc|cc|cc|cc}
\toprule
\multirow{2}{*}{Method} & \multicolumn{2}{c|}{$T=a$ / $S=m$} & \multicolumn{2}{c|}{$T=a$ / $S=y$} & \multicolumn{2}{c|}{$T=b$ / $S=m$} & \multicolumn{2}{c|}{$T=b$ / $S=y$} & \multicolumn{2}{c|}{$T=e$ / $S=m$} & \multicolumn{2}{c|}{$T=e$ / $S=y$} & \multicolumn{2}{c|}{$T=e$ \& $b$ / $S=m$} & \multicolumn{2}{c}{$T=a$ / $S=m$ \& $y$} \\
       & EO ($\downarrow$) & Acc. ($\uparrow$) & EO ($\downarrow$) & Acc. ($\uparrow$) & EO ($\downarrow$) & Acc. ($\uparrow$) & EO ($\downarrow$) & Acc. ($\uparrow$) & EO ($\downarrow$) & Acc. ($\uparrow$) & EO ($\downarrow$) & Acc. ($\uparrow$) & EO ($\downarrow$) & Acc. ($\uparrow$) & EO ($\downarrow$) & Acc. ($\uparrow$) \\
\hline \hline
CE~\cite{ce_loss_nips}     & 27.8 & 79.6 & 16.8 & 79.8 & 17.6 & 84.0 & 14.7 & 84.5 & 15.0 & 83.9 & 12.7 & 83.8 & 12.9 & 72.6 & 31.3 & 79.5 \\
GRL~\cite{raff2018reversal}    & 24.9 & 77.2 & 14.7 & 74.6 & 14.0 & 82.5 & 10.0 & 83.3 & 6.7  & 81.9 & 5.9  & 82.3 & 9.4  & 71.4 & 22.9 & 78.6 \\
LNL~\cite{kim2019notlearn}    & 21.8 & 79.9 & 13.7 & 74.3 & 10.7 & 82.3 & 6.8  & 82.3 & 5.0  & 81.6 & 3.3  & 80.3 & 7.4  & 70.8 & 20.7 & 77.7 \\
FD-VAE~\cite{Park2021disentanglement} & 15.1 & 76.9 & 14.8 & 77.5 & 11.2 & 81.6 & 6.7  & 81.7 & 5.7  & 82.6 & 6.2  & 84.0 & 8.2  & 70.2 & 19.9 & 78.0 \\
MFD~\cite{jung2021distillation}    & 7.4  & 78.0 & 14.9 & 80.0 & 7.3  & 78.0 & 5.4  & 78.0 & 8.7  & 79.0 & 5.2  & 78.0 & 9.0  & 70.0 & 19.4 & 76.1 \\
SupCon~\cite{supcon} & 30.5 & 80.5 & 21.7 & 80.1 & 20.7 & 84.6 & 16.9 & 84.4 & 20.8 & 84.3 & 10.8 & 84.0 & 12.5 & 72.7 & 24.4 & 81.7 \\
\midrule
FSCL (w/ group norm)~\cite{fscl}* & 6.5  & 79.1 & 12.4 & 79.1 & 4.7  & 82.9 & 4.8  & 84.1 & 3.0  & 83.4 & \textbf{1.6}  & 83.5 & 2.5  & 70.8 & 17.0 & 77.2 \\
\rowcolor{LightGreen}
\shasam (w/ LogDetCMI)   & 6.1  & 80.7 & 10.5 & 78.6 & 3.6  & 84.5 & 4.2  & 85.9 & 2.7  & 85.0 & 1.7  & 84.0 & 2.5  & 71.3 & 15.8 & 78.7 \\
\rowcolor{LightGreen}
\shasam (w/ FLCMI)   & \textbf{5.5}  & \textbf{81.3} & \textbf{9.9}  & \textbf{79.6} & \textbf{3.3}  & \textbf{84.7} & \textbf{3.9}  & \textbf{87.0} & \textbf{2.6}  & \textbf{85.8} & \textbf{1.6}  & \textbf{84.4} & \textbf{2.2}  & \textbf{71.8} & \textbf{14.6} & \textbf{79.5} \\
\midrule
FSCL (w/ ViT)~\cite{fscl}*        & 5.7  & 79.9 & 10.8 & 80.0 & 4.0  & 85.0 & 4.6  & 88.2 & 2.9  & 87.7 & 1.7  & 86.3 & 2.5  & 72.8 & 16.3 & 81.8 \\
TADet~\cite{Sudhakar2023MitigatingBI}  & 7.2  & 78.8 & 11.9 & 77.4 & 3.9 & 84.3 & 5.5 & 85.4 & 3.8 & 87.5 & 2.73 & 86.0 & 2.59 & 72.8 & 17.9 & 79.9 \\
FairViT~\cite{tian2024fairvit}* & 6.4  & 83.8 & 12.6 & \textbf{82.5} & 5.1  & \textbf{86.1} & 4.7  & 89.3 & 3.4  & 88.1 & 1.8  & \textbf{86.8} & 2.9  & \textbf{74.4} & 17.4 & 82.4 \\
\rowcolor{LightGreen}
\shasam (w/ FLCMI, ViT) & \textbf{5.3}  & \textbf{84.0} & \textbf{9.9}  & 82.3 & \textbf{3.0}  & 85.6 & \textbf{3.7}  & \textbf{89.2} & \textbf{2.8}  & \textbf{88.2} & \textbf{1.6}  & 86.8 & \textbf{2.2}  & 74.3 & 14.6 & \textbf{82.4} \\
\bottomrule
\end{tabular}}
\label{tab:results_celeba}
\end{table*}

\noindent (2) \textbf{UTKFace dataset}~\cite{utkface} comprises about 20,000 facial images with annotations for gender, age, and ethnicity. We set 'age' and 'ethnicity' as sensitive attributes and 'gender' as the target attribute. We transform 'age' and 'ethnicity' into binary attributes based on whether the age is under 35 or not, and whether the ethnicity is Caucasian or not, respectively~\cite{fscl, Shen2021ContrastiveLF}. In our constructed datasets, one sensitive group (e.g., Caucasian) contains $\alpha$ times more male data than female data, while the other sensitive group has the opposite gender ratio. We set $\alpha$ to 2, 3, and 4 to simulate different levels of bias. Unlike the training set, we organize completely balanced validation and test sets to ensure fair evaluation.\looseness-1

\noindent \textbf{Experimental Setup} We adopt a two-stage training strategy~\cite{supcon, fscl} to train the feature extractor $F$ and the classifier $Clf$ in \shasam. We follow \citet{fscl} and choose a ResNet-18~\cite{resnet} based architecture for $F$ and a vision-transformer based architecture~\cite{vit} for contrasting against FairViT~\cite{tian2024fairvit}.
Following the augmentation strategy described in~\cite{supcon}, we generate two cropped patches per image in the dataset and resize them to 128 $\times$ 128 pixels to feed as input to $F$. 
We train the encoder networks for 20 epochs during the representation learning phase (stage 1) optimized through $L_{\shasam}$ with a set batch size to 96 (16 anchors, 16 positives and 16 negatives in each iteration with 2 augmentations per image), an initial learning rate of 0.4, a cosine annealing scheduler and a temperature of 0.7. The latent space dimensions are set to 256 for the encoder and 128 for the projection network following the observations in \citet{supcon}.
Next, we freeze the feature extractor (from stage 1) and training $Clf$ (stage 2) with one hidden layer for 10 epochs using cross-entropy loss~\cite{ce_loss_nips}, a batch size of 128 and a constant learning rate of 0.1. 

\noindent \textbf{Metrics} Following most works in image classification we report the Top-1 Accuracy of predicting the target attribute $T$. For estimating the fairness aspect of \shasam\ we report the Equalized Odds (EO)~\cite{hardt2016eo} which focuses on both the discrepancy between the True Positive Rate (TPR) and False Positive Rate (FPR) among sensitive attribute groups.

A low value of EO indicates non-discriminative/ insensitive response of the model to the sensitive attributes indicating higher fairness in predictions.
Although measures like Demographic Parity (DP)~\cite{demo_parity} and Equal Opportunity (EOpp) exists in literature but \cite{fscl} points out the sensitivity to deliberate mis-classifications and discounting unfairness in negative classes as major pitfalls in DP and EOP respectively. The experimental results are aggregated over several independent runs on 4 NVIDIA A6000 GPUs with additional details in the Appendix.

\subsection{Contrasting Against Baselines}
We compare the performance of \shasam\ against several SoTA benchmarks as baselines - vanilla CE~\cite{ce_loss_nips}, GRL~\cite{raff2018reversal}, LNL~\cite{kim2019notlearn}, FD-VAE~\cite{Park2021disentanglement}, MFD~\cite{jung2021distillation}, FSCL~\cite{fscl} and FairViT~\cite{tian2024fairvit} and report the top-1 accuracy (Acc.) and Equalized Odds (EO) in \cref{tab:results_celeba}.
Although CE, LNL and GRL improve the learning of decision boundaries between target classes in CelebA characterized by gains in Top-1 Accuracy, all of them demonstrate significantly high EO indicating unfair predictions. 
Compared to SoTA method FSCL, \shasam\ (with FLCMI) observes up to 2.7 point reduction in EO with a max. of 2.95\% boost in accuracy when the backbone architecture was chosen as ResNet-18~\cite{resnet}. 
Replacing the ResNet-18 based encoder with a ViT~\cite{vit} encoder we continue to observe up to 1.9 points reduction in EO with a 4.1\% boost in performance.
Further contrasting against FairViT which proposes a novel backbone architecture insensitive to sensitive attributes, \shasam\ achieves up to 2.8 points reduction in EO without compromising accuracy gains. 
Our FL based learning strategy indicated as \shasam\ w/ FLCMI shows the highest improvement in fairness with competitive boosts in Top-1 accuracy over SoTA methods.\looseness-1

\begin{figure}
        \centering
        \includegraphics[width=\columnwidth]{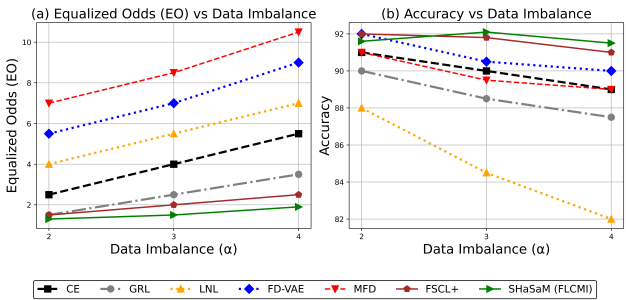}
        \caption{\textbf{Results of \shasam\ on UTKFace dataset} measuring (a) Equalized Odds and (b) Top-1 Acc. under varying inter-group imbalance ($\alpha$). The target and sensitive attributes are set to \textit{gender} and $\textit{ethnicity}$ respectively following setup in \citet{fscl}.}
        \label{fig:res_utkface}
        \vspace{-4ex}
\end{figure}

Further, we compare the resilience of \shasam\ in overcoming imbalance between sensitive attribute groups. To this end we vary the imbalance ratio $\alpha$ between sensitive attribute groups in the UTKFace dataset among 2,3,4 and report the top-1 accuracy and EO through \cref{fig:res_utkface}(b) and \cref{fig:res_utkface}(a). We show that our \shasam\ w/  FLCMI approach shows improvements in both EO and accuracy under varying degrees of imbalance. 
This clearly demonstrates the effectiveness of \shasam\ in learning fair representations from the target classes without biasing on the features from sensitive classes.\looseness-1

\subsection{Ablation Study}
\noindent \textbf{Components in \shasam}
In this section we contrast the two main components of \shasam\ - \shasam-MINE and \shasam-LEARN against baseline method FSCL to demonstrate the effectiveness of our hard sample modeling formulation. At first, we consider random sampling of anchors, positives and negatives, a common practice in the field~\cite{fscl, supcon, Shen2021ContrastiveLF} while varying the learning strategy. We observe in \cref{tab:abl_components_celeba} that \shasam-LEARN provides significant gains (reduction here) in EO over baselines while demonstrating marginal gains in accuracy. 
Next, we keep the learning strategy constant (\shasam-LEARN) and vary the hard sample mining between random and \shasam-MINE. 
Enforcing a large decision boundary between target groups while learning features invariant to sensitive attributes through the joint selection and learning formulation in \shasam\ demonstrates the best accuracy and EO over baselines.

\begin{figure*}[th]
        \centering
        \includegraphics[width=0.95\textwidth]{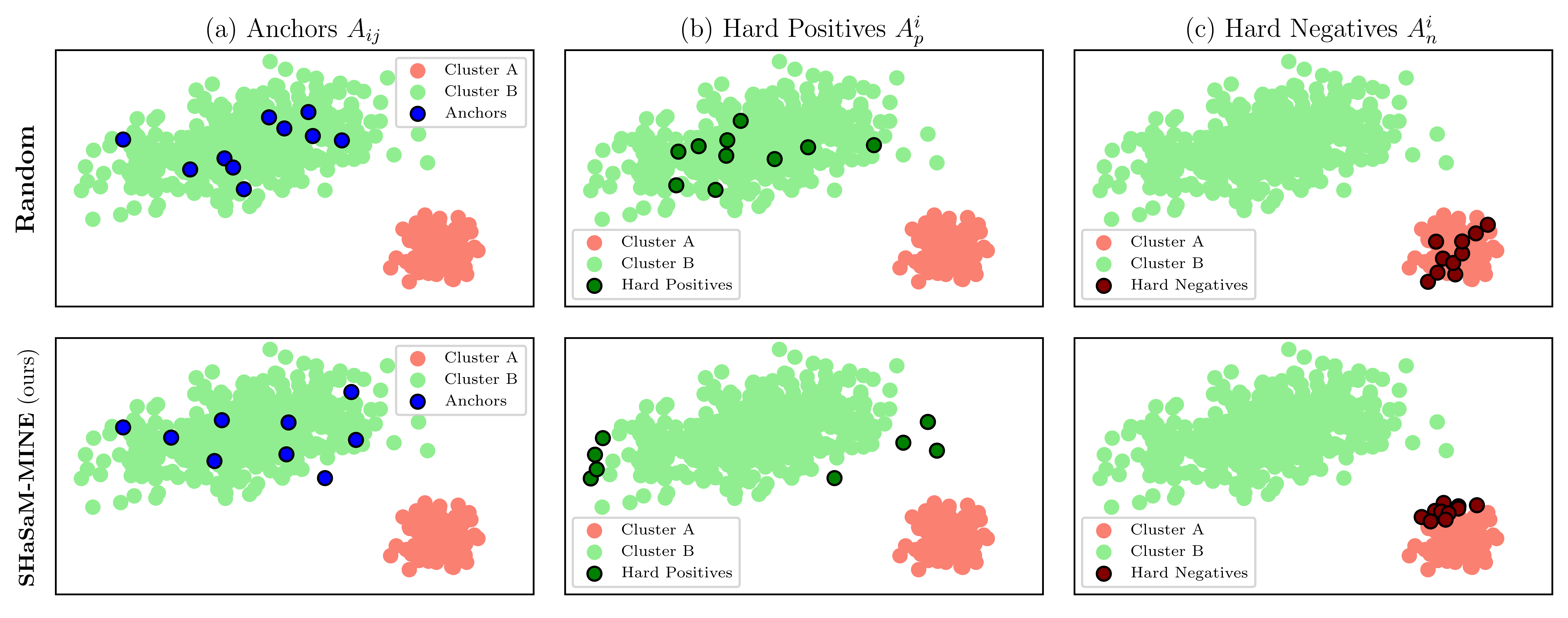}
        \caption{\textbf{Contrasting Random and \shasam-MINE selection strategies} on a synthetic two-cluster imbalanced dataset to identify (a) Anchors, (b) Hard Positives and (c) Hard Negatives, showing the effectiveness of \shasam\ in modeling the decision boundary between target attributes. The dataset generation and sample selection in performed under the same seed.}
        \label{fig:shasam_mine_res}
        \vspace{-2ex}
\end{figure*}

\begin{table}[tb]
\caption{\textbf{Ablation on components in \shasam} measured in terms of Top-1 Accuracy (Acc.) and equalized odds (EO) by varying the target $T$ and sensitive attributes $S$ on CelebA dataset. Here, FL indicates FLCMI and LD indicates LogDetCMI functions.\looseness-1}
\centering
\resizebox{\columnwidth}{!}{
\begin{tabular}{c|c|cc|cc}
\toprule
Selection & Learning & \multicolumn{2}{c|}{$T=a$ / $S=m$} & \multicolumn{2}{c}{$T=a$ / $S=y$} \\
Strategy  & Objective & EO ($\downarrow$) & Acc. ($\uparrow$) & EO ($\downarrow$) & Acc. ($\uparrow$)   \\
\hline \hline
Random       & SupCon~\cite{supcon}  &  30.5 & 80.5 &	21.7 & 80.1 \\
Random       & FSCL~\cite{fscl}      &  6.5  & 79.1 &   12.4 & 79.1 \\
Random       & \shasam-LEARN (LD)    &  6.2  & 79.4 &   11.2 & 79.1\\ 
Random       & \shasam-LEARN (FL)    &  5.7  & 80.3 &   9.0  & 79.4\\ 
\midrule
\shasam-MINE (FL) & SupCon~\cite{supcon}  &  24.6 & \textbf{81.5} &	14.3 & \textbf{81.0} \\
\shasam-MINE (FL) & FSCL~\cite{fscl}      &  5.9 & 79.0 &   9.9  & 79.3\\ 
\shasam-MINE (LD) & \shasam-LEARN (FL)    &  6.1 & 78.8 &   10.6 & 77.8 \\
\shasam-MINE (LD) & \shasam-LEARN (LD)    &  6.1 & 80.7 &   9.9 & 78.6 \\
\shasam-MINE (FL) & \shasam-LEARN (FL)    &  \textbf{5.6} & 81.3 &  \textbf{9.88} & 79.58 \\ 
\bottomrule
\end{tabular}}
\label{tab:abl_components_celeba}
\vspace{-2ex}
\end{table}

\noindent \textbf{Sample Selection in \shasam-MINE}
We qualify the effectiveness of \shasam-MINE by conducting experiments on a 2-cluster (denoted as $A$ and $B$ respectively) imbalanced (imabalance ratio $\alpha$=5) synthetic dataset representing target attributes. We contrast \shasam-MINE against random selection which is a common practice in most recent works~\cite{fscl, tian2024fairvit, Shen2021ContrastiveLF}. \cref{fig:shasam_mine_res} depicts the selected anchors, positives and negatives by both \textit{Random} and \textit{\shasam-MINE} strategies with the target class selected as $A$. Our \shasam-MINE selects diverse anchors from the target attribute class while selecting hard positives (samples located farthest to the anchor within the target cluster) and hard negatives (samples located closest to the anchors outside the target cluster) which random sampling fails to guarantee.
Minimizing $L_{\shasam}$ with hard positives and negatives (which lie on the cluster boundary) mined through \shasam-MINE ensures sufficient inter-attribute separation, a critical feature in fair representation learning. We provide more details in the Appendix.

\begin{figure}
        \centering
        \includegraphics[width=\columnwidth]{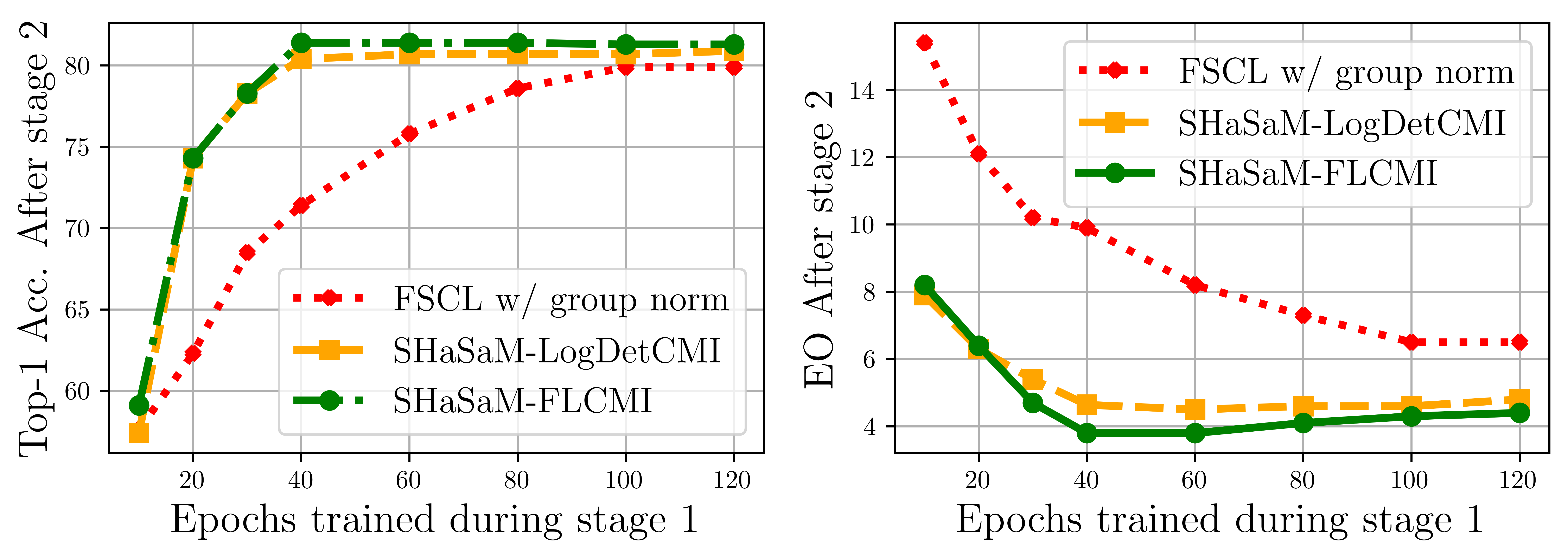}
        \caption{The set formulation in \shasam\ learns discriminative representations \textbf{within fewer training epochs} (in stage 1) in CelebA for \textit{male} and \textit{attractiveness} as target and sensitive attributes.}
        \label{fig:shasam_convergence}
        \vspace{-4ex}
\end{figure}

\noindent \textbf{Convergence of \shasam-LEARN}
Through \cref{fig:shasam_convergence} we compare the number of training epochs in stage 1 required to learn discriminative feature representations for various instances of \shasam\ against baseline FSCL (with group-normalization). Keeping the experimental settings same as \cref{sec:experiments} and batch size at 96 across methods we train $F$ for specific number of epochs in the range of [10, 100] in stage 1 while training $Clf$ in stage 2 for 10 epochs. The reported accuracy and EO values are reported on the CelebA dataset with \textit{gender} (male or not) and \textit{attractiveness} as target and sensitive attribute respectively. We show that \textbf{\shasam\ (with FLCMI based learning objective) achieves similar performance to FSCL within fewer epochs} in stage 1. Nevertheless, the joint selection and learning strategy in \shasam\ potentially increases per iteration wall-clock time, further discussed in the Appendix.\looseness-1

\section{Conclusion}
In conclusion, we introduce \textbf{\shasam}, a novel combinatorial framework that addresses the critical challenge of fairness in facial attribute recognition by framing fairness-driven representation learning as a submodular hard-sample mining problem. Through its two-stage approach—\shasam-MINE and \shasam-LEARN—our framework leverages submodular functions to select balanced sets of hard-positive and hard-negative samples, mitigating attribute imbalance and preventing sensitive attributes from influencing the learned representations. By minimizing the Conditional Mutual Information between target and sensitive attributes, \shasam-LEARN enhances fairness while preserving downstream performance. Empirical results on popular benchmarks demonstrate that \shasam\ improves fairness (with up to 2.7 point improvement in Equalized Odds) within fewer training epochs compared to existing methods. This work highlights the potential of combinatorial viewpoint in advancing fairness in facial attribute recognition.

\section*{Acknowledgements}
We gratefully thank anonymous reviewers for their valuable comments. We would also like to extend our gratitude to our fellow researchers from the CARAML lab at UT Dallas for their suggestions. This work is supported by the National Science Foundation under Grant Numbers IIS-2106937, a gift from Google Research, an Amazon Research Award, and the Adobe Data Science Research award. Any opinions, findings, and conclusions or recommendations expressed in this material are those of the authors and do not necessarily reflect the views of the National Science Foundation, Google or Adobe.

{
    \small
    \bibliographystyle{ieeenat_fullname}
    \bibliography{references}
}

% \section{Appendix}
% \input{sections/07-appendix}
% Start the appendix
\appendix
\appendixpage
\startcontents[sections]
\printcontents[sections]{l}{1}{\setcounter{tocdepth}{2}}

\section{Notation}
Following the problem definition in the main paper we introduce the notations used in \cref{tab:notations} throughout the paper.
\begin{table*}[t]
      \caption{Collection of notations used in the paper.}
      \centering
      \resizebox{0.9\textwidth}{!}{\begin{tabular}{ c | c }
            \toprule
           \textbf{Symbol}  & \textbf{Description} \\
            \midrule
            $\mathcal{V}$ & The Ground set, here refers to the mini-batch at each iteration. \\
            $T_i$ & The target attribute set $T_i$, $\forall i \in |\mathcal{V}|$. \\
            $S_j$ & The sensitive attribute set $S_j$, $\forall j \in |\mathcal{V}|$. \\
            $A_{ij}$ & Anchor set with examples from target attribute $T_i$ and sensitive attribute $S_j$. \\
            $A_p^i$ & Hard-Positives with examples from target attribute $T_i$ and sensitive attribute $\overline{S_j}$. \\
            $A_n^i$ & Hard-Negatives with examples from target attribute $\mathcal{V} \setminus T_i$ and sensitive attribute $S_j$. \\
            $F(x, \theta)$ & Neural Network used as feature extractor. \\
            $Clf(.,.)$ & Multi-Layer Perceptron as classifier. In our case a two layer network. \\
            $\theta$ & Parameters of the feature extractor. \\
            $S_{A,B}(\theta)$ & Cross-Similarity between sets $A, B \in \mathcal{V}$. \\
            $S_A(\theta)$ & Self-Similarity between samples in set $A \in \mathcal{T}$. \\
            $f(A)$ & Submodular Information function over a set $A$. \\
            $I_f(A; Q)$ & Submodular Mutual Information function between sets $A$ and $Q$. \\
            $H_f(A|Q)$ & Submodular Conditional Gain function between sets $A$ and $Q$. \\
            $I_f(A; Q| P)$ & Submodular Conditional Mutual Information function between a target set $A$, query set $Q$ and private set $P$. \\
            $L_{\shasam}(\theta)$ & Loss value computed over all target and sensitive attribute pairs. \\
            $N_f(A_{ij})$ & Normalization constant approximated to $3 |A_{ij}|$. \\
            $EO$ & Equalized Odds. \\
            \bottomrule
      \end{tabular}}
      \label{tab:notations}      
\end{table*}

\section{Additional Related Work and Preliminaries}
\subsection{Contrastive Learning}
In the realm of supervised learning, conventional models utilizing Cross-Entropy (CE) loss~\citep{ce_loss_nips} often grapple with challenges posed by class imbalance and noisy labels. To address these issues, metric learning techniques~\citep{arcface, cosface, opl, msloss} aim to learn distance-based~\citep{triplet} or similarity-based~\citep{arcface, cosface} metrics, fostering orthogonality within the feature space~\citep{opl} and bolstering class-specific feature discrimination. Contrastive learning, rooted in noise contrastive estimation~\citep{nce}, has become a cornerstone in self-supervised learning~\citep{simclr, moco, mocov2}, where label information is unavailable during training. In supervised contexts, SupCon~\citep{supcon} emphasizes forming feature clusters rather than merely aligning features to predefined centroids. For instance, Triplet loss~\citep{triplet} differentiates one positive and one negative pair, whereas N-pairs~\citep{n_pairs} loss incorporates multiple negative pairs, and SupCon extends this by leveraging multiple positive and negative pairs. Lifted-Structure loss~\citep{lsl} sharpens focus by contrasting positives against the hardest negatives, and SupCon exhibits similarities with Soft-Nearest Neighbors loss~\citep{snn}, which maximizes inter-class entanglements. Despite the significant achievements of these methods, they predominantly rely on pairwise similarity metrics, which may not inherently facilitate the formation of disjoint clusters.\looseness-1

\subsection{Submodularity (Cont. from Section 3.2)}
As discussed in Sec. 3.2 of the main paper, submodular functions have been recognized to model notions of cooperation~\cite{submod_cooperation}, diversity~\cite{submod_diversity}, representation~\cite{prism} and coverage~\cite{vid_sum_2019}. 
Following the combinatorial formulation in Sec.3.1 of the main paper we define the ground set $\mathcal{V} = \{T_1, T_2, \cdots T_N\} = \{S_1, S_2, \cdots S_K\}$ and explore four different categories of submodular information functions in our work, namely - 

\noindent \textbf{(1)} \textit{Submodular Total Information} ($S_f$) which measures the total information contained in each set~\cite{fujishige}, expressed as $S_f(T_1, T_2, \dots, T_N)$ as in \cref{eq:sim}. Maximizing $S_f$ over a set $T_i$ models diversity~\cite{submod_diversity} while minimizing $S_f$ models cooperation~\cite{submod_cooperation}.\looseness-1
\begin{align}
    S_f(T_1, T_2, \dots, T_N) = \overset{N}{\underset{i=1}{\sum}}f(T_i)
    \label{eq:sim}
\end{align}

\noindent \textbf{(2)} \textit{Submodular Mutual Information} ($I_f$) which models the shared information between two sets~\cite{prism} which serves as a measure of \textit{similarity/cooperation} between them, expressed through \cref{eq:smi}.\looseness-1
\begin{align}
    I_f(T_i; T_j) = f(T_i) + f(T_j) - f(T_i \cup T_j), \forall i,j \in |\mathcal{V}|
    \label{eq:smi}
\end{align}

\noindent \textbf{(3)} \textit{Submodular Conditional Gain} ($H_f$) which models the gain in information when a set $T_j$ is added to $T_i$. $H_f$ models the notion of \textit{dissimilarity} between sets and can be expressed in \cref{eq:scg}.
\begin{align}
\begin{split}
    H_f(T_i| T_j) &= f(T_i \cup T_j) - f(T_j) \\ 
                  &= f(T_i) - I_f(T_i; T_j) \text{ , }  \forall i,j \in |\mathcal{V}|
\end{split}
\label{eq:scg}    
\end{align}

\noindent \textbf{(4)} \textit{Submodular Conditional Mutual Information} ($I_f$) which jointly models the mutual similarity between two sets $T_i$ and $T_j$ and their collective dissimilarity to a conditioning set $C$ as: 
\begin{align}
\begin{split}
    I_f(T_i; T_j| C) &= f(T_i \cup C) + f(T_j \cup C) \\
    &- f(T_i \cup T_j \cup C) - f(C) \\ 
    I_f(T_i; T_j | C) &= I_f(T_i \cup C; T_j) - I_f(T_j; C) \\ 
    &= H_f(T_i | C) + H_f(T_j | C) \\
    &- H_f(T_i \cup T_j | C) \text{ , }  \forall i,j \in |\mathcal{V}|
\end{split}
\label{eq:cmi}
\end{align}

Note, that the above formulations can also be reformulated by considering the sensitive attribute $S_i$ instead of $T_i$ , $\mathcal{V} = \cup_{i = 1}^N T_i = \cup_{j = 1}^{K} S_j$. 
Given a submodular function $f$ (can alternatively be $I_f$ or $H_f$) tasks like selection~\cite{jain2023efficient,killamsetty_automata} and summarization~\cite{kaushal2019framework, vid_sum_2019} have been modeled as a discrete optimization problem to identify a summarized set of examples $A \subseteq \mathcal{V}$ via submodular maximization under a cardinality constraint ($|A| \leq k$), i.e. $max_{A \subseteq \mathcal{V}, |A| \leq k} f(A)$. This can be fairly approximated with a $(1 - e^{-1})$ constant factor guarantee~\cite{Nemhauser1978} using greedy optimization techniques~\cite{Mirzasoleiman2015lazierthanlazy} as shown in \cref{alg:greedy}.
Extending the definition of submodular functions to continuous optimization space \citet{score} have proposed a set of novel family of learning objectives which minimize total information and total correlation among sets in $D_{train}$ using continuous optimization techniques like SGD. These objectives have been shown to be significantly more robust to large imbalance demonstrated in real-world tasks like longtail recognition~\cite{score} and few-shot learning~\cite{smile}.\looseness-1

\begin{algorithm}
\label{alg:greedy}
\caption{Greedy Submodular Maximization as in \citet{Nemhauser1978}.}
\begin{algorithmic}[1]
\Require Submodular function $f : 2^{\mathcal{V}} \rightarrow \mathbb{R}$, cardinality constraint $\mathtt{k}$
\Ensure Set $A \subseteq \mathcal{V}$ maximizing $f(A)$ under cardinality constraint $\mathtt{k}$
\State Initialize an empty set $A \leftarrow \emptyset$
\For{$j = 1$ to $\mathtt{k}$}
    \State $e \leftarrow \underset{v \in \mathcal{V} \setminus A}{\texttt{argmax}} (f(A \cup \{v\}) - f(A))$
    \State $A \leftarrow A \cup \{e\}$
\EndFor
\State \Return $A$
\end{algorithmic}
\end{algorithm}

\section{Additional Explanation to Fig. 1}
In this section we elaborate the steps of operation of \shasam\ which is condensed and depicted in Fig. 1 of the main paper. 
At first, given a ground set (a large labeled pool) $\mathcal{V}$ we would like to mine an anchor set $A_{12}$. Anchor sets are supposed to contain examples sharing the same target and sensitive attributes $T_1$ and $S_2$ as shown in Fig. 1. In our example $T_1$ refers to males and $\overline{T_1}$ refers to non-males. Similarly $S_2$ refers to males wearing sunglasses and $\overline{S_2}$ refers to people (irrespective of gender) that do not wear sunglasses. Thus, $A_{12}$ contains examples of males wearing eyeglasses.
Once we have the anchor set, we now use the formulation described in Sec. 3.3 we mine $A_p^1$ which resembles the hard-positives containing examples of males which do not wear eyeglasses. Similarly we now mine hard-negatives $A_n^1$ which contain examples of females (or $\overline{T_1}$) who wear eyeglasses as shown in Fig. 1(a). This is the operation of \shasam-MINE.

Once we have mined $A_{12}$, $A_p^1$ and $A_n^1$ we now would like to learn embedding separation between $A_{12}$ and $An^1$ while bringing $A_p^1$ closer to $A_{12}$. As shown in Fig. 1(b) we apply \shasam-LEARN which is a loss function derived from Submodular Conditional Mutual Information (SCMI) as shown in eq. 3 of the main paper. Following the discussion in Sec. 3.5 SCMI based learning objectives jointly model anchor-hard-negative separation while modeling anchor-hard-positive cooperation when minimized used SGD. With sufficient training \shasam-LEARN results in an embedding space which learns the decision boundary between males and non-males without biasing on the sensitive attribute - wearing sunglasses as shown in Fig. 1(b).

\section{Additional Implementation Details}
The implementation details are largely elucidated in Sec. 4 of the main paper but we add more details below and at \url{https://anaymajee.me/assets/project_pages/shasam.html}.

\subsection{Settings for Selection in \shasam-MINE}
As discussed in Algorithm 1 of the main paper we employ three different submodular functions to sample the anchors $A_{ij}$, hard-positives $A_p^i$ and hard-negatives $A_n^i$. 
We run the selection at every iteration of the training process with 16 anchors, 16 hard-positives and 16 hard-negatives selected in each iteration. Each image further gets augmented into two views which results in a batch size of 96 as discussed in Sec. 4 of the main paper. 
Following our problem formulation, at each iteration we randomly select a target attribute $T_i$ and a sensitive attribute $S_j$ where $i,j \in |\mathcal{V}|$. All anchors in $A_{ij}$ share the same target and sensitive attribute, thus mined from $T_i \cap S_j$, while the hard-positives share the same target attribute but different sensitive attribute labels from $T_i \cap \overline{S_j}$. Finally, negatives are mined from $\overline{T_i} \cap S_j$ and share orthogonal target attributes while retaining the same sensitive attribute. This formulation follows \citet{fscl} and results in selection of samples at the cluster boundary.

Since the ground set resembles the complete dataset at each iteration, selection of $A_{ij}$, $A_p^i$ and $A_n^i$ becomes computationally inefficient (large computational cost).
To mitigate this situation, we randomly select a subset of the dataset at each epoch and use it for model training. This strategy draws inspiration from \citet{okanovic2024repeated} and significantly improves training speeds without compromising on performance.

\subsection{Training and Evaluation in \shasam-LEARN}
As discussed in Sec. 3 of the main paper, \shasam-MINE and \shasam-LEARN are sequentially invoked at each iteration for training the feature extractor $F$ on dataset $D^{train}$. Following the experimental setup in Sec. 4 of the main paper, each image in both CelebA and UTKFace datasets are resized to 128 $\times$ 128 pixels with two complementary augmentations among - RandomCrop, Grayscaling, ColorJitters etc. similar to \citet{supcon}.
For the implementations of GRL~\cite{raff2018reversal}, LNL~\cite{kim2019notlearn}, FD-VAE~\cite{Park2021disentanglement}, MFD~\cite{jung2021distillation}, FSCL~\cite{fscl} we follow the implementation in \citet{fscl} and report the average Top-1 Accuracy (Acc.) and Equalized Odds (EO) over three random seeds. Particularly in case of FSCL, we adopt their two-stage training strategy to train a ResNet-18 based feature extractor on the train split for 100 epochs with a 0.1 initial learning rate and a cosine annealing scheduler. On the contrary, our \shasam\ approach is trained with a higher initial learning rate of 0.4 with a cosine annealing scheduler for $\sim$ 20 epochs during stage 1. For stage 2 both FSCL and \shasam\ are trained for 10 epochs, keeping the feature extractor frozen.

For FairViT~\cite{tian2024fairvit} we adopt a ViT~\cite{vit} based feature extractor with fairness driven modifications as suggested in the original paper. We perform a single stage training till convergence. Unlike using the first 80 individuals from the CelebA dataset as in the original implementation, we use all individuals available in CelebA for reproduction of the results. For fair comparisons, we replace the feature extractor in \shasam\ (ResNet-18) with a vanilla ViT architecture and conduct two sets of experiments, one following FSCL (denoted as FSCL w/ViT in Table 2) and the other adopting the \shasam\ (denoted as \shasam\ w/ FLCMI, ViT in Table 2) with Facility-Location Conditional Mutual Information (FLCMI) based learning objective.

\subsection{Experiments on Synthetic Dataset}
To characterize the selection strategy discussed in Sec. 3.4 (\shasam-MINE) we conduct experiments on synthetic datasets by varying (1) \textit{imbalance} and (2) \textit{Feature Similarity} (cluster overlap). 
To this end we introduce a 2-dimensional 2-cluster setting as shown in Fig. 4 of the main paper. For varying imbalance we ablate on the imbalance ratio $\alpha$ from 1 in the balanced setting and 5 in the imbalanced setting. Fig. 4 (main paper) depicts the imbalanced setting with $\alpha = 5$. The abundant cluster $A$ consists of 500 examples with high intra-group variance while the rare class has 100 examples with low intra-group variance. The selection budget was set to 10 examples for each anchor, hard-positive and hard-negative sets. To contrast against \textit{Random} sampling (commonly used strategy in \cite{supcon, fscl, Shen2021ContrastiveLF}) we keep the seed constant for underlying libraries.
Between the imbalanced settings Fig. 4 (main paper) and \cref{fig:shasam_mine_res_minority} (supplementary), we highlight the difference in selected samples by altering the target attribute between the majority and minority class. 
For the balanced setting (1), $\alpha$ is set to 1 with each cluster containing 200 examples. This is depicted in \cref{fig:shasam_mine_res_balanced}. The variance of the clusters indicated through the spread of points in the feature space is also kept constant between cluster $A$ and $B$.
For varying the (2) feature similarity we reduce the separation between two clusters so that some overlap exists between them by reducing the 2D distance between the cluster centroids. The imbalance is kept constant at $\alpha = 5$ and the majority class is considered as the target class,similar to setting in Fig. 4 of the main paper. This setting is depicted through \cref{fig:shasam_mine_res_overlap} and discussed in detail in \cref{app:synthetic_data_experiment}.

\section{Gradients through $L_{\shasam}(\theta)$}
In this section we provide proof for our proposed formulation in Sec. 3.2, eq. (2) which states that the gradient through $L_{\shasam}(\theta) \approx \nabla L_{\shasam}(A_{ij}, A_p^i, A_n^i, \theta)$ which is the gradient over the combinatorial loss alone. 

\begin{proof}
    At first we can consider the definition of $L_{\shasam}(\theta)$ as a product of three functions $X(\theta)$, $Y(\theta)$ and $Z(\theta)$ as shown below.

    \begin{align}
\begin{split}
   L_{\shasam}(\theta) = &\sum_{\forall i,j \in |\mathcal{V}|} 
   \underbrace{\texttt{softmax}(H_f(.|A_{ij}), A_p^i, T_i \cap \overline{S_j}))}_\text{$X(\theta)$} \\
     &\times \underbrace{\texttt{softmax}(I_f(.;A_{ij}), A_n^i, \mathcal{V} \setminus T_i \cap S_j))}_\text{$Y(\theta)$} \\
   & \times \underbrace{L_{\shasam}(A_{ij}, A_p^i, A_n^i ; \theta)}_\text{$Z(\theta)$} \\
   L_{\shasam}(\theta) = & \sum_{\forall i,j \in |\mathcal{V}|} X(\theta) \times Y(\theta) \times Z(\theta)
\end{split}
   \label{eq:formulation}
\end{align}
Given this simplified form, we calculate the gradients through $L_{\shasam}(\theta)$ following the chain rule as -
\begin{align}
\begin{split}
    \frac{\partial L_{\shasam}(\theta)}{\partial \theta} = & \frac{\partial X(\theta)}{\partial \theta} \times Y(\theta) \times Z(\theta) \\
    & + X(\theta) \times \frac{\partial Y(\theta)}{\partial \theta} \times Z(\theta) \\
    & + X(\theta) \times Y(\theta) \times \frac{\partial Z(\theta)}{\partial \theta} 
\end{split}
\end{align}

We know that, $X(\theta)$ and $Y(\theta)$ are hard-positive and hard-negative miners and are approximated as the $\texttt{softmax}$ over the selected subset $A$ from $Q$ given a selection function $F(.)$. Lets call this $\sigma$ -
\begin{align}
\sigma_i (\theta) = \texttt{softmax}(F(.; \theta), A, Q) = \frac{\exp(F_i(A; \theta))}{\sum_{j} \exp(F_j(A; \theta))}
\end{align}
Irrespective of the choice of the selection function $F(.)$ the gradient over $\sigma$ can be written as - 
\begin{align}
    \frac{\partial \sigma_i(\theta)}{\partial F_i(A)} = 
    \begin{cases}
    \sigma_i(\theta) (1 - \sigma_i(\theta)) & \text{if } i = k, \\
    -\sigma_i(\theta) \sigma_k(\theta) & \text{if } i \neq k.
    \end{cases}
\end{align}

In the first case, when $i == k$ the softmax function evaluates to 1 and occurs when $\sigma_i(\theta)$ is approximately the \texttt{argmax}. Similarly, for the second case, $i \neq k$ occurs when $\sigma_i(\theta)$ does not approximate the argmax. In this case $\sigma_i(\theta)$ evaluates to 0. Thus, in both cases the gradients through $X(\theta)$ and $Y(\theta)$ approximate to 0. 
This reduces the gradient calculation for $L_{\shasam}(\theta)$ as -
\begin{align}
    \begin{split}
        \frac{\partial L_{\shasam}(\theta)}{\partial \theta} = X(\theta) \times Y(\theta) \times \frac{\partial Z(\theta)}{\partial \theta}
    \end{split}
\end{align}
Since $Z(\theta)$ represents the loss function $L_{\shasam}(A_{ij}, A_p^i, A_n^i, \theta)$ and thus:
\begin{align}
    \begin{split}
        % \frac{\partial L_{\shasam}(\theta)}{\partial \theta} =&  C. \frac{\partial L_{\shasam}(A_{ij}, A_p^i, A_n^i, \theta)}{\partial \theta} \\
        \nabla_{\theta} L_{\shasam}(\theta) \approx &  \nabla_{\theta} L_{\shasam}(A_{ij}, A_p^i, A_n^i, \theta)
    \end{split}
\end{align}
Where C is a constant, $C = X(\theta). Y(\theta)$. This proves our consideration in the formulation in Sec. 3.2 and allows \shasam\ to utilize a mixture of discrete and continuous optimization problems to model learning of fair representations as a submodular hard sample mining.
\end{proof}

\section{Derivation of Instances of \shasam}
We define the loss function in \shasam\ as the conditional mutual information between the mined anchors $A_{ij}$, hard-positives $A_p^i$ and hard-negatives $A_n^i$ as shown in eq. (3) of the main paper. We summarize it here for readability as \cref{eq:cmi_repeat}.
\begin{align}
   L_{\shasam}(\theta) &= \sum_{\forall i,j \in |\mathcal{V}|} \frac{1}{N_f(A_{ij})} I_f\left(A_{ij}; A_n^i | A_p^i; \theta\right)
   \label{eq:cmi_repeat}
\end{align}
In this section we derive two important instances of $L_{\shasam}$ based on the choice of the submodular function $f(A, \theta)$ over set $A$ as summarized in Table 1 of the main paper.

\subsection{\shasam-FLCMI}
Given a dataset $\mathcal{V}$ and the Facility-Location (FL) submodular function $f(A) = \sum_{i \in \mathcal{V}} \underset{j \in A}{\max} S_{ij}$ over a set $A$, we derive a combinatorial loss $L_{\shasam}$ on the mined anchors $A_{ij}$, hard-positives $A_p^i$ and hard-negatives $A_n^i$ based on the Submodular Conditional Mutual Information function $I_f(A_{ij}; A_n^i|A_p^i)$ as shown in \cref{eq:shasam_flcmi}.

\begin{align}
\begin{split}
    L_{\shasam}(\theta) =& \underset{i,j \in |\mathcal{V}|}{\sum} \frac{1}{3|A_{ij}|} \max\Biggl(\min\Biggl(\underset{a \in A_{ij}}{\max} S_{ia}, \\
    &\underset{n \in A_n^i}{\max} S_{in}\Biggr) - \underset{p \in A_p^i}{\max} S_{ip}, 0 \Biggr)
\end{split}
\label{eq:shasam_flcmi}
\end{align}

\begin{proof}
    From the definition of CMI based on conditional gain as shown in \cref{eq:cmi},
    \begin{align}
    \begin{split}
        I_f(A_{ij}; A_n^i|A_p^i) =& H_f(A_{ij} | A_p^i) + H_f(A_{ij} | A_n^i) \\
        &- H_f(A_{ij} \cup A_n^i | A_p^i) \\
        =& f(A_{ij}) - I_f(A_{ij}; A_p^i) \\
        &+ f(A_{ij}) - I_f(A_{ij}; A_n^i) \\
        &- f(A_{ij} \cup A_n^i) + I_f(A_{ij} \cup A_n^i; A_p^i) 
    \end{split}
    \label{eq:if_as_hf_flcmi}
    \end{align}

    Now, we separate each term containing $I_f$ and expand them based on the definition of SMI in \cref{eq:smi} - 
    \begin{align}
        \begin{split}
        I_f(A_{ij}; A_p^i) =& f(A_{ij}) + f(A_p^i) - f(A_{ij}\cup A_p^i) \\    
        =& \sum_{i,j \in \mathcal{V}} \underset{a \in A_{ij}}{\max} S_{ia} + \sum_{i,j \in \mathcal{V}} \underset{p \in A_p^i}{\max} S_{ip} \\
        &- \sum_{i,j \in \mathcal{V}} \underset{k \in A_{ij} \cup A_p^i}{\max} S_{ik} \\
        =& \sum_{i,j \in \mathcal{V}} \underset{a \in A_{ij}}{\max} S_{ia} +  \underset{p \in A_p^i}{\max} S_{ip} \\
        &- \max\left(\underset{k \in A_{ij}}{\max} S_{ik}, \underset{k \in A_p^i}{\max} S_{ik}\right) \\
        =& \sum_{i,j \in \mathcal{V}} \min\left(\underset{k \in A_{ij}}{\max} S_{ik}, \underset{k \in A_p^i}{\max} S_{ik}\right)
        \end{split}        
    \end{align}
    Similarly we can calculate $I_f(A_{ij}; A_n^i)$ and $I_f(A_{ij} \cup A_n^i; A_p^i)$ as - 
    \begin{align}
        I_f(A_{ij}; A_n^i) = \sum_{i,j \in \mathcal{V}} \min\left(\underset{k \in A_{ij}}{\max} S_{ik}, \underset{k \in A_n^i}{\max} S_{ik}\right)
    \end{align}
    \begin{align}
        I_f(A_{ij} \cup A_n^i; A_p^i) = \sum_{i,j \in \mathcal{V}} \min\left(\underset{k \in A_{ij} \cup A_n^i}{\max} S_{ik}, \underset{k \in A_p^i}{\max} S_{ik}\right)
    \end{align}

    Substituting these expressions in \cref{eq:if_as_hf_flcmi} and simplifying we get - 
    \begin{align}
        \begin{split}
        I_f(A_{ij}; A_n^i|A_p^i) =& H_f(A_{ij} | A_p^i) + H_f(A_{ij} | A_n^i) \\
        &- H_f(A_{ij} \cup A_n^i | A_p^i) \\
        =&\sum_{i,j \in \mathcal{V}} \max\left(0, \underbrace{\underset{k \in A_{ij}}{\max} S_{ik}}_\text{p} - \underbrace{\underset{k \in A_p^i}{\max} S_{ik}}_\text{r}\right) \\
        &+ \max\left(0, \underbrace{\underset{k \in A_n^i}{\max} S_{ik}}_\text{q} - \underbrace{\underset{k \in A_p^i}{\max} S_{ik}}_\text{r}\right) \\
        &- \max\left(0, \underset{k \in A_{ij} \cup A_n^i}{\max} S_{ik} - \underset{k \in A_p^i}{\max} S_{ik}\right)
        \end{split}
    \end{align}
    This follows the expression $\max(p - r, 0) + \max(q - r, 0) - \max(\max(p,q) - r, 0)$. Which evaluates to -
    \begin{align}        
        \begin{cases}
            \max(q - r, 0) & \text{if } p > q, \\
            \max(p - r, 0) & \text{if } p < q
        \end{cases}
    \end{align}
    Thus we can simplify the expression of $I_f(A_{ij}; A_n^i|A_p^i)$ as -
    \begin{align}
    \begin{split}
         I_f(A_{ij}; A_n^i|A_p^i) =& \underset{i,j \in \mathcal{V}}{\sum}  \max\Biggl(\min\Biggl(\underset{k \in A_{ij}}{\max} S_{ik}, \\
        &\underset{k \in A_n^i}{\max} S_{ik}\Biggr) - \underset{k \in A_p^i}{\max} S_{ik}, 0 \Biggr)
    \end{split}
    \end{align}
    Substituting this in the expression of $L_{\shasam}(\theta)$ we get -
    \begin{align}
    \begin{split}
        L_{\shasam}(\theta) =& \underset{i,j \in |\mathcal{V}|}{\sum} \frac{1}{3|A_{ij}|} \max\Biggl(\min\Biggl(\underset{k \in A_{ij}}{\max} S_{ik}, \\
        &\underset{k \in A_n^i}{\max} S_{ik}\Biggr) - \underset{k \in A_p^i}{\max} S_{ik}, 0 \Biggr) \text{        ... Hence proved.}
    \end{split}
    \end{align}
\end{proof}

\subsection{\shasam-LogDetCMI}
Given a dataset $\mathcal{V}$ and the Log-Determinant (LogDet) submodular function $f(A) = \text{logdet}(S_A)$ over a set $A$, we derive a combinatorial loss $L_{\shasam}$ on the mined anchors $A_{ij}$, hard positives $A_p^i$ and hard negatives $A_n^i$ based on the Submodular Conditional Mutual Information function $I_f(A_{ij}; A_n^i|A_p^i)$ as shown in \cref{eq:shasam_logdetcmi}.
\begin{equation}
    \resizebox{\columnwidth}{!}{
    $L_{\shasam}(\theta) = \sum_{i,j \in |\mathcal{V}|} \frac{1}{3|A_{ij}|} \log \frac{\det \left( I - S_{A_n^i}^{-1} S_{{A_n^i},{A_p^i}} S_{A_p^i}^{-1} S_{{A_n^i},{A_p^i}}^T \right)}{\det \left( I - S_{A_{ij} \cup {A_n^i}}^{-1} S_{A_{ij} \cup {A_n^i},{A_p^i}} S_{A_p^i}^{-1} S_{A_{ij} \cup {A_n^i},{A_p^i}}^T \right)}$}
    \label{eq:shasam_logdetcmi}
\end{equation}

\begin{proof}
    From the definition of CMI based on Mutual Information as described in \cref{eq:cmi}, we get - 
    \begin{align}
    \begin{split}
        I_f(A_{ij}; A_n^i | A_p^i) =& I_f(A_{ij} \cup A_p^i ; A_n^i) - I_f(A_n^i ; A_p^i) \\
        =& f(A_{ij} \cup A_p^i) + f(A_n^i) - f(A_{ij} \cup A_p^i \cup A_n^i) \\
        &- f(A_n^i) + f(A_p^i) - f(A_n^i \cup A_p^i)
    \end{split}
    \end{align}
    Given the definition of LogDet over a set $A$, $f(A) = \text{logdet}(S_A)$, we substitute this in the above equation to get the following - 
    \begin{align}
    \begin{split}
        I_f(A_{ij}; A_n^i | A_p^i) =& \text{logdet}(S_{A_{ij} \cup A_p^i}) + \text{logdet}(S_{A_n^i}) \\
        &- \text{logdet}(S_{(A_{ij} \cup A_p^i) \cup A_n^i}) \\
        &- \text{logdet}(S_{A_n^i}) + \text{logdet}(S_{A_p^i}) \\
        &- \text{logdet}(S_{A_n^i \cup A_p^i}) \\
        =& \log\Biggl(\frac{\text{det}(S_{A_{ij} \cup A_p^i}). \text{det}(S_{A_n^i})}{\text{det}(S_{A_{ij} \cup A_p^i \cup A_n^i})}\Biggr)\\
        &- \log\Biggl(\frac{\text{det}(S_{A_n^i}). \text{det}(S_{A_p^i})}{\text{det}(S_{A_n^i \cup A_p^i})}\Biggr)
    \end{split}
    \end{align}
    From Schur's complement we know that $det(S_{A \cup B}) = det(S_A) det(S_{A \cup B} \setminus S_A)$ and $S_{A \cup B} \setminus S_A = S_B - S_{A,B}^T S_A^{-1}S_{A,B}$, where $S_{A,B}$ refers to the cross-similarities between sets $A$ and $B$ while $S_A$ and $S_B$ represent the corresponding self-similarities. We use this to simplify the expression of $I_f$ above as follows - 
    \begin{align}
        \begin{split}
            I_f(A_{ij}; A_n^i | A_p^i) =& \log\Biggl(\frac{\text{det}(A_n^i)}{\text{det}(S_{A_{ij} \cup A_p^i \cup A_n^i} \setminus S_{A_{ij} \cup A_p^i})}\Biggr)\\
            &- \log\Biggl(\frac{\text{det}(S_{A_p^i})}{\text{det}(S_{A_n^i \cup A_p^i} \setminus S_{A_n^i})}\Biggr) \\
            =& \log\text{det}\Biggl(\frac{S_{A_n^i \cup A_p^i} \setminus S_{A_n^i}}{S_{A_p^i}}\Biggr) \\
            &- \log\text{det}\Biggl(\frac{S_{A_{ij} \cup A_p^i \cup A_n^i} \setminus S_{A_{ij} \cup A_p^i}}{S_{A_n^i}}\Biggr) \\
            I_f(A_{ij}; A_n^i | A_p^i) =& \text{logdet}\Biggl(\frac{S_{A_p^i} - S_{A_n^i,A_p^i}^T S_{A_n^i}^{-1}S_{A_n^i,A_p^i}}{S_{A_p^i}}\Biggr) \\
            - \text{logdet}\Biggl(&\frac{S_{A_n^i} - S_{A_{ij} \cup A_p^i, A_n^i}^T S_{A_{ij} \cup A_p^i}^{-1}S_{A_{ij} \cup A_p^i,A_n^i}}{S_{A_n^i}}\Biggr) \\
            =& \text{logdet}\left(I - S_{A_p^i}^{-1}S_{A_n^i,A_p^i}^T S_{A_n^i}^{-1}S_{A_n^i,A_p^i}\right) \\
            - \text{logdet}\Biggl(I - &S_{A_n^i}^{-1}S_{A_{ij} \cup A_p^i,A_n^i}^T S_{A_{ij} \cup A_p^i}^{-1}S_{A_{ij} \cup A_p^i,A_n^i}\Biggr) \\            
        \end{split}
    \end{align}
    Following simple logarithmic principles we can further simplify this expression as -
    \begin{equation}
        \resizebox{\columnwidth}{!}{
        $I_f(A_{ij}; A_n^i | A_p^i) = \log \frac{\det \left( I - S_{A_n^i}^{-1} S_{{A_n^i},{A_p^i}} S_{A_p^i}^{-1} S_{{A_n^i},{A_p^i}}^T \right)}{\det \left( I - S_{A_{ij} \cup {A_p^i}}^{-1} S_{A_{ij} \cup {A_p^i},{A_n^i}} S_{A_n^i}^{-1} S_{A_{ij} \cup {A_p^i},{A_n^i}}^T \right)}$}
    \end{equation}
    Substituting this in the equation of $L_{\shasam}(\theta)$ in \cref{eq:cmi_repeat} we get the loss function for \shasam-LogDetCMI as shown below.
    \begin{equation}
        \resizebox{\columnwidth}{!}{
        $L_{\shasam}(\theta) = \sum_{i,j \in |\mathcal{V}|} \frac{1}{3|A_{ij}|} \log \frac{\det \left( I - S_{A_n^i}^{-1} S_{{A_n^i},{A_p^i}} S_{A_p^i}^{-1} S_{{A_n^i},{A_p^i}}^T \right)}{\det \left( I - S_{A_{ij} \cup {A_p^i}}^{-1} S_{A_{ij} \cup {A_p^i},{A_n^i}} S_{A_n^i}^{-1} S_{A_{ij} \cup {A_p^i},{A_n^i}}^T \right)}$}
    \end{equation}
\end{proof}

\section{Additional Fairness Metrics (Contd. from Sec. 4)}
\label{app:additional_metrics}
To provide a comprehensive evaluation of our model, we assess both its predictive performance and its adherence to fairness principles. 
In addition the reported metrics in Sec. 4 of the main paper we also evaluate our model on additional metrics that have been used in SoTA approaches.

For performance, we utilize \textbf{Balanced Accuracy} (BA), a metric particularly effective for datasets with imbalanced class distributions. Rather than a simple accuracy score, BA calculates the average of the true positive and true negative rates across each subgroup, offering a more nuanced view of a model's effectiveness and ensuring that high performance on a majority group does not obscure poor performance on a minority group. 

For fairness, we employ two widely accepted metrics. The first, \textbf{Demographic Parity} (DP), measures whether the rate of positive predictions is consistent across different sensitive groups. A model satisfies DP if its decisions are statistically independent of an individual's group membership, with a value approaching zero indicating greater fairness. 

The second metric, \textbf{Equalized Opportunity} (EOpp), enforces a more stringent fairness condition by requiring that the model's true positive rate is the same for all sensitive groups. This ensures the model is equally effective at correctly identifying positive instances, regardless of group identity. For both DP and EOpp, scores closer to zero signify a more equitable model.

\begin{table}[tb]
\caption{\textbf{Ablation on selection budget in \shasam} measured in terms of Top-1 Accuracy (Acc.) and equalized odds (EO) by varying the budget $\mathtt{k}$ on two settings in CelebA dataset. The learning objective was kept constant at \shasam-LEARN w/ FLCMI and the selection strategy in \shasam-MINE is set to FLCMI.\looseness-1}
\centering
\resizebox{\columnwidth}{!}{
\begin{tabular}{c|c|cc|cc}
\toprule
Selection & Budget & \multicolumn{2}{c|}{$T=a$ / $S=m$} & \multicolumn{2}{c}{$T=a$ / $S=y$} \\
Strategy & $\mathtt{k}$ & EO ($\downarrow$) & Acc. ($\uparrow$) & EO ($\downarrow$) & Acc. ($\uparrow$) \\
\midrule
FSCL~\cite{fscl} & -  & 6.5 & 79.1 & 12.4 & 79.1 \\
\shasam          & 8  & 7.0 & 79.9 & 10.2 & 77.2 \\
\shasam          & 16 & 5.6 & 81.3 &  9.9  & 79.58 \\
\shasam          & 24 & 5.5 & 81.08 & 10.1  & 79.44 \\
\bottomrule
\end{tabular}}
\label{tab:abl_budget}
\vspace{-2ex}
\end{table}

\section{Additional Results}

\begin{figure}
        \centering
        \includegraphics[width=\columnwidth]{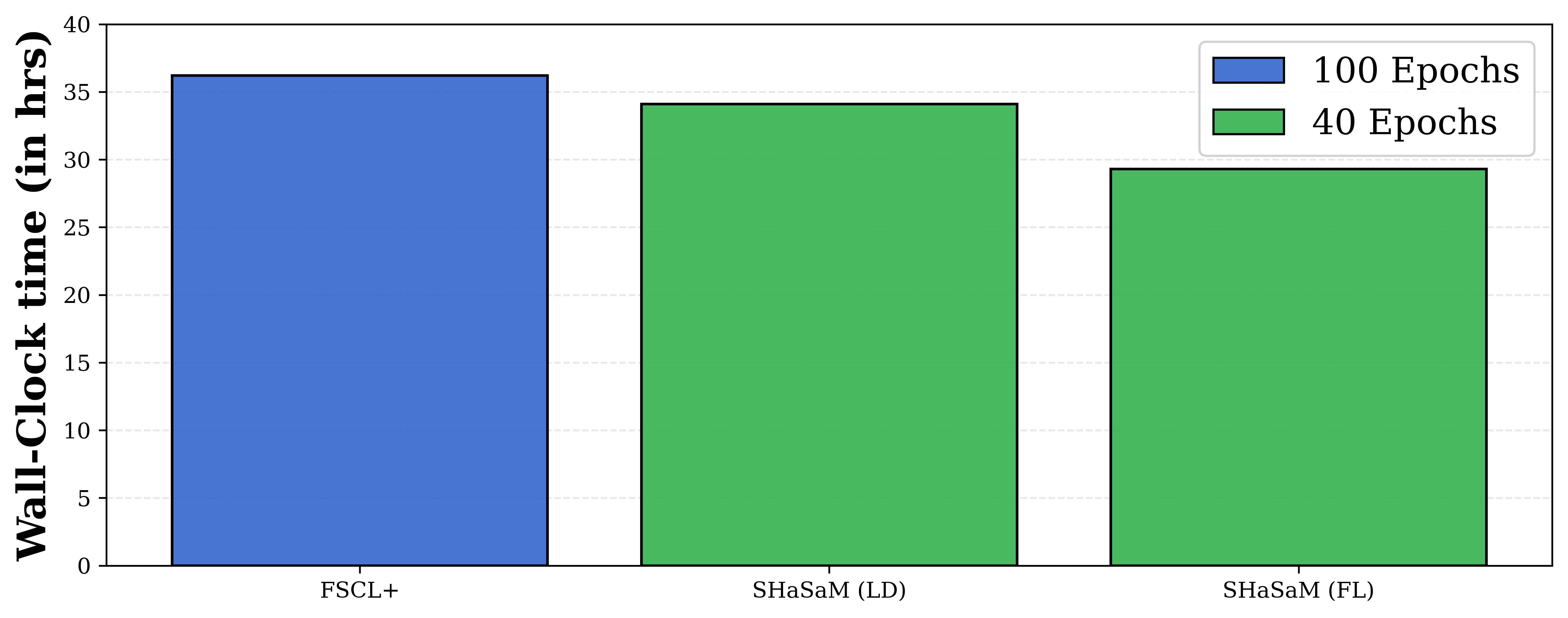}
        \caption{\textbf{Ablation on Computational Cost} measured in terms of wall clock time averaged over three settings in CelebA dataset. The submodular function in the learning objective and selection strategy in \shasam\ was kept constant.}
        \label{tab:abl_cost_wall_clock}
        \vspace{-4ex}
\end{figure}

\begin{table*}[ht]
\centering
\caption{A comparison of different methods across three distinct tasks. The tasks involve predicting an attribute ($T$) while maintaining fairness with respect to a sensitive attribute ($S$). We report Accuracy (Acc.), Balanced Accuracy (BA), Equalized Opportunity (EOpp.), and Demographic Parity (DP). Higher ACC and BA are better, while lower EO and DP are better. The best result in each column is highlighted in bold.}
\label{tab:fairvit_fairness_cont}
% Use resizebox to make the table fit within the page width
\resizebox{\textwidth}{!}{%
\begin{tabular}{l|cccc|cccc|cccc}
\hline
\multirow{2}{*}{method} & \multicolumn{4}{c|}{$T=a$ / $S=m$} & \multicolumn{4}{c|}{$T=s$ / $S=m$} & \multicolumn{4}{c}{$T=a$ / $S=br$} \\ \cline{2-13}
 & Acc.$_{\%}$ & BA$_{\%}$ & EOpp.$_{e-2}$ & DP$_{e-1}$ & Acc.$_{\%}$ & BA$_{\%}$ & EOpp.$_{e-2}$ & DP$_{e-1}$ & Acc.$_{\%}$ & BA$_{\%}$ & EOpp.$_{e-2}$ & DP$_{e-1}$ \\ \hline\hline
Vanilla & 74.01 & 72.36 & 14.43 & 3.245 & 88.42 & 88.85 & 4.91 & 1.489 & 76.48 & 74.55 & 3.61 & 1.896 \\
TADeT-MMD~\cite{ramaswamy2021latentdebiasing} & 79.89 & 73.85 & 7.10 & 3.693 & 92.51 & 93.03 & 2.48 & 1.290 & 77.97 & 75.64 & 2.27 & 1.491 \\
TADeT~\cite{ramaswamy2021latentdebiasing} & 78.73 & 74.52 & 3.11 & 3.116 & 90.05 & 90.68 & 4.86 & 1.443 & 78.49 & 77.42 & 3.78 & 1.057 \\
FSCL~\cite{fscl} & 79.09 & 74.76 & 1.78 & 3.004 & 89.37 & 90.08 & 1.76 & 1.344 & 78.85 & 78.06 & 2.65 & 0.989 \\
FSCL+~\cite{fscl} & 77.26 & 73.42 & \textbf{0.79} & 2.604 & 88.83 & 89.02 & 1.20 & 1.263 & 78.02 & 77.37 & 1.79 & 0.834 \\
FairViT~\cite{tian2024fairvit} & 83.80 & 79.96 & 1.15 & 2.837 & \textbf{94.27} & 94.12 & 1.52 & 1.205 & \textbf{82.52} & \textbf{81.56} & 2.10 & \textbf{0.701} \\ 
\rowcolor{LightGreen}
\shasam (w/ FLCMI, ViT) (ours) & \textbf{84.01} & \textbf{80.16} & 0.76 & \textbf{2.582} & 92.87 & \textbf{94.57} & \textbf{1.17} & \textbf{1.193} & 80.70 & 79.42 & \textbf{2.06} & 0.873 \\ \hline
\end{tabular}%
}
\end{table*}

\subsection{Additional Results from Metrics in \cref{app:additional_metrics}}
Although Equalized Odds is the most commonly adopted fairness metric in literature there exists metrics such as Demographic Parity (DP), Equalized Opportunity (EOpp) and Balanced Accuracy (BA) that are studied in the context of fair facial attribute recognition. 
In addition to Equalized Odds (EO) in Tab. 2 (main paper) we present results from the aforementioned metrics by closely following the exact benchmark in FairViT~\cite{tian2024fairvit} in \cref{tab:fairvit_fairness_cont}. Similar to FairViT we also adopt three distinct settings by varying $T$ and $S$ - $T$ = attractiveness ($a$) / $S$ = gender (male, $m$), $T$ = expression (smiling, $s$) / $S$ = gender (male, $m$), $T$ = attractiveness ($a$) / $S$ = hair-color (brown, $br$).
Similar to our observations in Tab. 2 (main paper) we see that \shasam\ (with FLCMI as the instance for \shasam-MINE and \shasam-LEARN) achieves competitive accuracy (indicated as Acc.) to FairViT without the requirement to train the Vision Transformer (ViT) architecture from scratch.
Finally, the self-balancing property as elucidated in \cite{score} ensures that majority classes do not bias the decisions made by \shasam. This is reflected in the boost in performance measured through BA across all downstream tasks.\looseness-1

\begin{figure}
        \centering
        \includegraphics[width=\columnwidth]{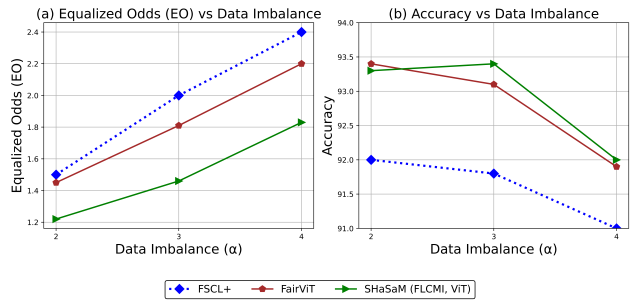}
        \caption{\textbf{Results of \shasam\ on UTKFace dataset on ViT based benchmark} (FairViT) measuring (a) Equalized Odds and (b) Top-1 Acc. under varying inter-group imbalance ($\alpha$). The target and sensitive attributes are set to \textit{gender} and $\textit{ethnicity}$ respectively following setup in \citet{fscl}.}
        \label{fig:res_utkface_vit}
        \vspace{-4ex}
\end{figure}

\subsection{Comparison against FairViT on UTKFace Dataset}
In continuation to the results presented in Sec. 4.1 on UTKFace dataset we include experiments contrasting the performance of \shasam\ against FairViT~\cite{tian2024fairvit}. For fair comparisons we replace the resnet based backbone in \shasam\ with a ViT based backbone and depicts the results in \cref{fig:res_utkface_vit} under similar imbalanced settings discussed in Sec. 4.1 and report the top-1 accuracy and EO through \cref{fig:res_utkface_vit}(b) and \cref{fig:res_utkface_vit}(a). Similar to Sec. 4.1 we show that our \shasam\ w/  FLCMI approach shows improvements in both EO and accuracy under varying degrees of imbalance. 

\subsection{Ablation on the selection Budget}
As discussed in Sec. 4 of the main paper, \shasam\ learns from a set of mined anchors, hard-positives and hard-negatives.
Since the selection of exemplars in each of these sets is performed through submodular optimization under the knapsack constraint~\cite{Nemhauser1978}, a fixed budget $\mathtt{k}$ is established for each set.
For simplicity we keep this budget constant across three sets discussed above such that in each iteration there are exactly $\mathtt{k}$ anchors, $\mathtt{k}$ hard-positives and $\mathtt{k}$ hard-negatives resulting in a total batch size of $3\mathtt{k}$ (excluding augmentations).
From the results tabulated in \cref{tab:abl_budget} a pattern emerges wherein a higher budget (more examples) benefit both fairness and accuracy metrics. However, its interesting to note that the performance saturates beyond $\mathtt{k} = 16$. Due to this (alongside compute limitations) we adopt a selection budget of 16 in all our experiments.
Note, we were unable to execute experiments with higher selection budgets in \cref{tab:abl_budget} due to compute limitations.

\subsection{Ablation: Choice of Combinatorial Function in \shasam}
\label{abl:choice_comb_func}
As discussed in Sec. 3.4 and 3.5 the choice of submodular function $f$ induces various combinatorial properties encoded in \shasam-MINE and \shasam-LEARN. In Tab. 3 of the main paper we vary the submodular function between LD (indicating LogDetCMI) and FL (indicating FLCMI) functions and report the performance both in-terms of EO and accuracy (indicated as Acc. in Tab. 3). At first, we show that FL based selection and learning functions demonstrate improved performance and fairness. This is because Facility-Location (the submodular function in these instances) models representation in contrast to diversity (modeled by LogDet), mining representative anchors, positives and negatives in \shasam-MINE while learning representative features in \shasam-LEARN.

\subsection{Ablation: Compute Cost and Wall Clock time}
We point out that introduction of the learning formulation in \shasam\ does not add any additional parameters to the model. Nevertheless, particular instances of \shasam\ like LogDetCMI requires computation of large matrices which scale with the increase in batch size requiring large compute infrastructure. We have discussed this in the limitations of our paper in \cref{sec:limitations}. 

Additionally, we also compare the wall-clock time requirements of \shasam\ against SoTA methods in \cref{tab:abl_cost_wall_clock}. 
For each setting in \cref{tab:abl_cost_wall_clock} we calculate the average time in phase 1 (phase 2 remains largely unchanged) across three distinct settings in CelebA.
Our observations closely follow the discussion in Sec. 4.2 (main paper) which shows that adopting a combinatorial approach in \shasam\ requires fewer training epochs in stage 1 (100 epochs in FSCL to $\sim$40 epochs in \shasam). However, the selection of hard positives and negatives do add a computation overhead which reflects in the wall-clock times shown in \cref{tab:abl_cost_wall_clock}. This has been listed as a limitation in \cref{sec:limitations}.

\begin{figure*}[th]
        \centering
        \includegraphics[width=\textwidth]{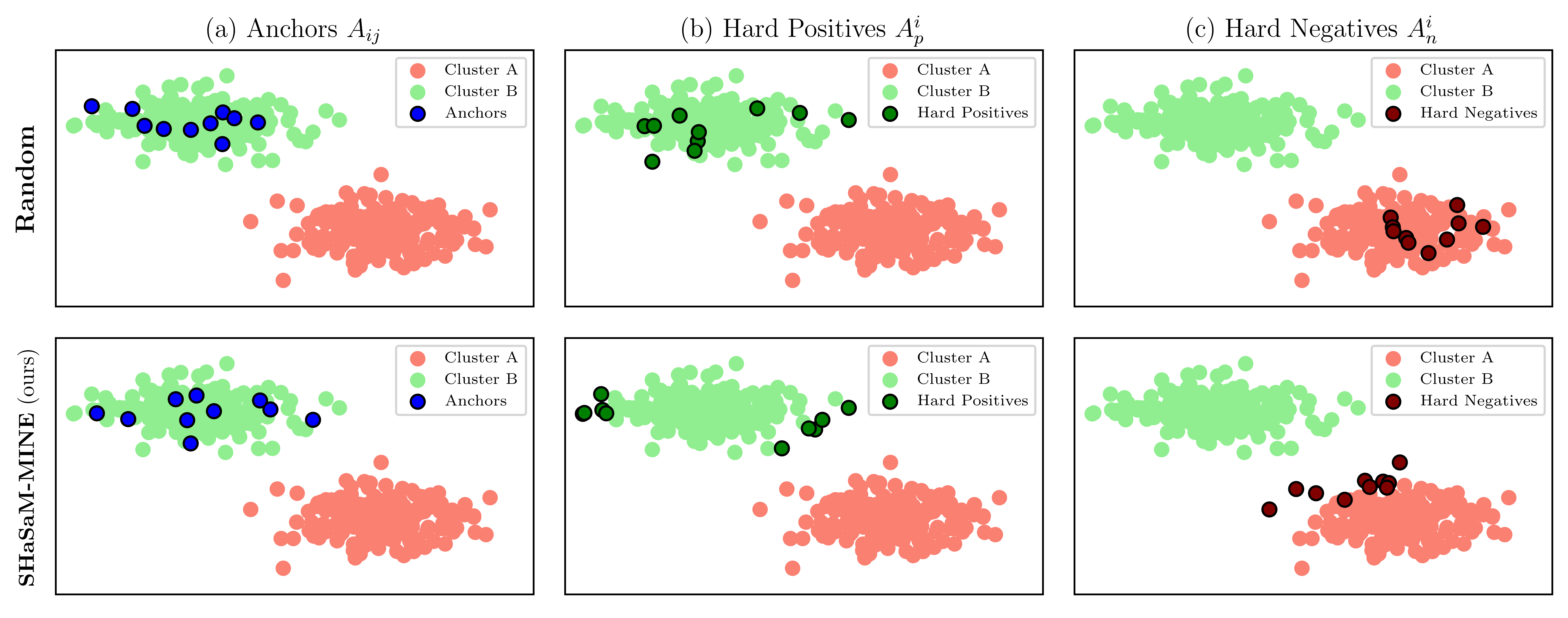}
        \caption{\textbf{Contrasting Random and \shasam-MINE selection strategies} on a synthetic two-cluster imbalanced dataset to identify (a) Anchors, (b) Hard Positives and (c) Hard Negatives, in the \textit{balanced} setting. The dataset generation and sample selection in performed under the same seed.}
        \label{fig:shasam_mine_res_balanced}
        \vspace{-2ex}
\end{figure*}

\subsection{Characterization of \shasam-MINE on Synthetic Data}
\label{app:synthetic_data_experiment}
To characterize the effectiveness of the introduced \shasam-MINE we simulate three different scenarios. These include variation in sample sizes between target attributes inducing \textit{imbalance} and simulating \textit{feature overlap} between groups inducing inter-group bias.
Our goal is to show that minimizing $L_{\shasam}$ on the mined anchors, hard positives and negatives facilitate the learning of strong decision boundaries between target attributes while ensuring learning of compact feature clusters (minimize intra-group variance) for each target attribute label without biasing on the sensitive attribute labels. 

\begin{figure*}[th]
        \centering
        \includegraphics[width=\textwidth]{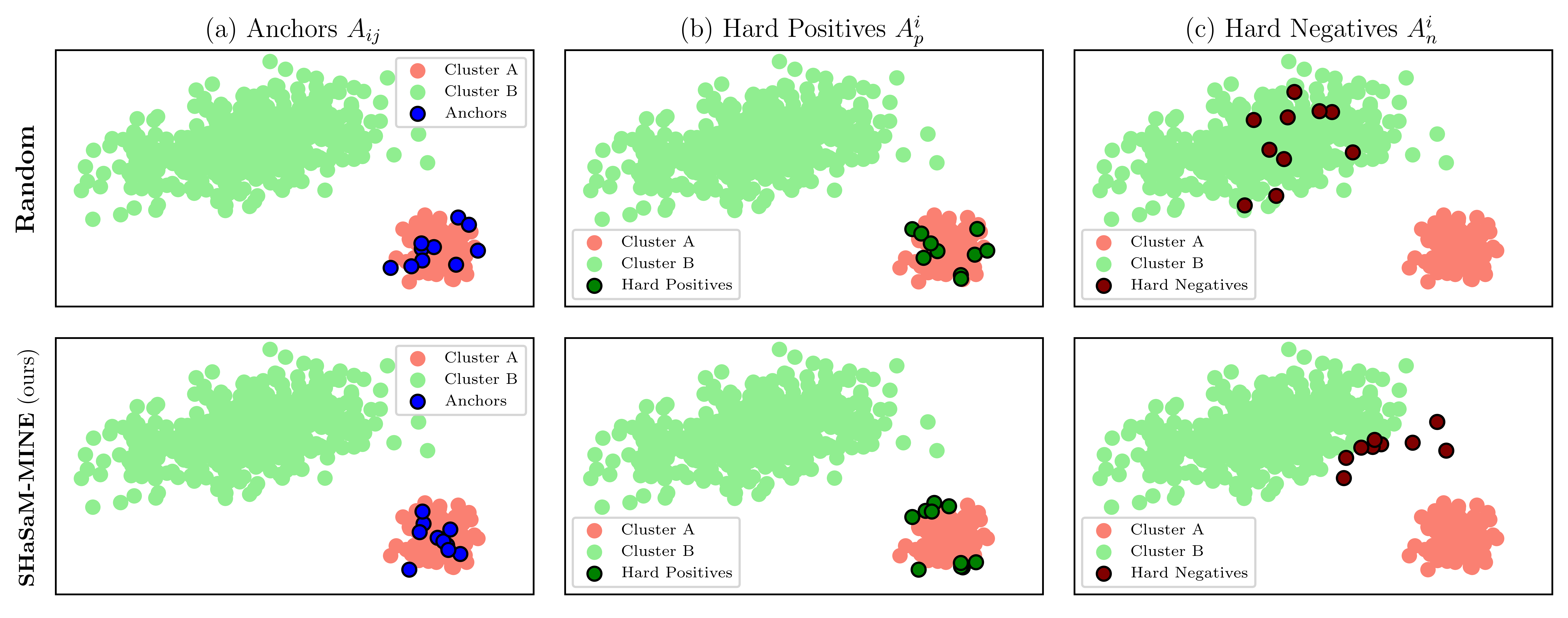}
        \caption{\textbf{Contrasting Random and \shasam-MINE selection strategies} on a synthetic two-cluster imbalanced dataset with target attribute as the minority (rare) class. \shasam-MINE identifies (a) Anchors, (b) Hard Positives and (c) Hard Negatives, showing the effectiveness of \shasam\ in modeling the decision boundary between target attributes. The dataset generation and sample selection in performed under the same seed.}
        \label{fig:shasam_mine_res_minority}
        \vspace{-2ex}
\end{figure*}

\noindent \textbf{Balanced Settings}
In this case both clusters $A$ and $B$ as shown in \cref{fig:shasam_mine_res_balanced} have equal number of samples with distinguishable decision boundary between them.
In contrast to \textit{Random} selection which is the most widely used technique, \shasam-MINE selects diverse anchors representing the complete target set (in this case cluster $B$). Additionally, our approach selects hard-positives which lie at the cluster boundary of $B$. Minimizing the separation between these hard-positives and anchors result in minimization of intra-group variance encouraging the model to be unbiased to the sensitive attributes.
Lastly, hard-negatives in $A$ also lie at the cluster boundary between $A$ and $B$, resulting in increased decision boundary when the separation between hard-positives and and negatives are maximized through $L_{\shasam}$.

\begin{figure*}[th]
        \centering
        \includegraphics[width=\textwidth]{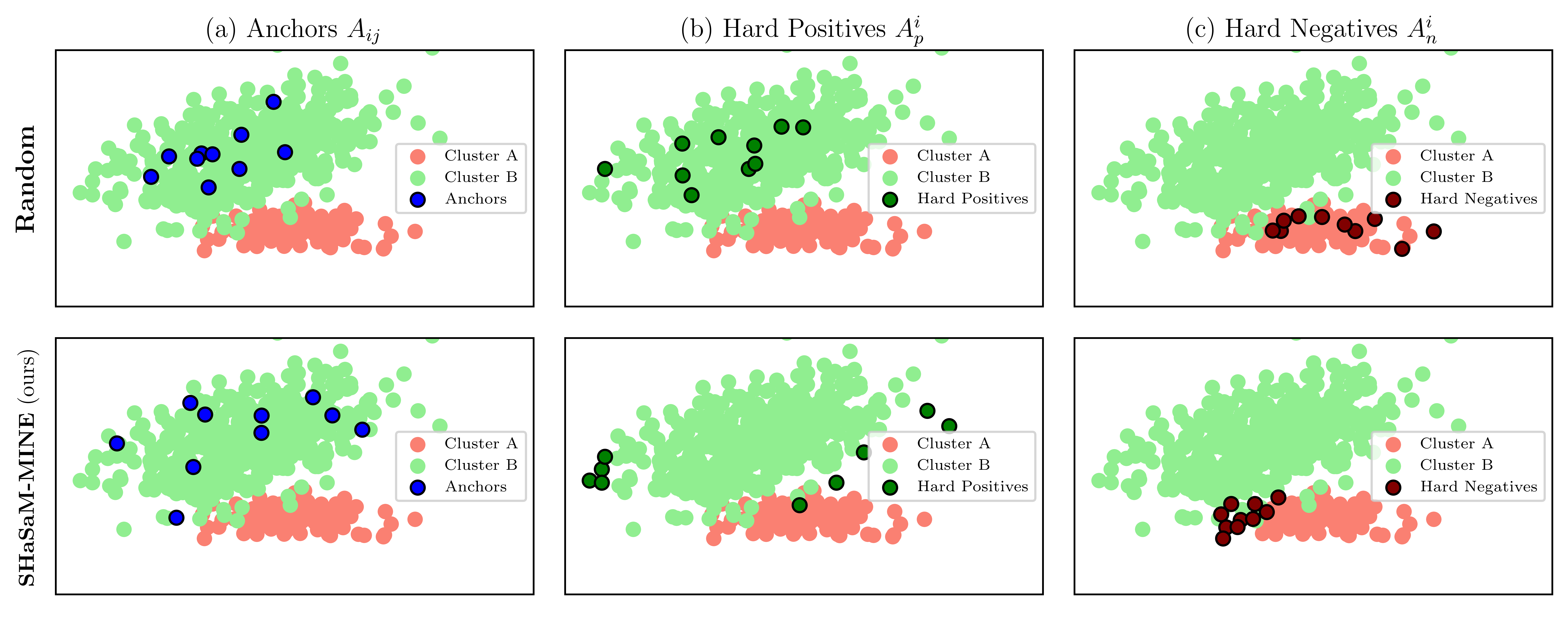}
        \caption{\textbf{Contrasting Random and \shasam-MINE selection strategies} on a synthetic two-cluster imbalanced dataset to identify (a) Anchors, (b) Hard Positives and (c) Hard Negatives, when large feature similarity exists between clusters (overlapping clusters). The dataset generation and sample selection in performed under the same seed.}
        \label{fig:shasam_mine_res_overlap}
        \vspace{-2ex}
\end{figure*}

\noindent \textbf{Imbalanced Settings}
In this case clusters $A$ and $B$ as shown in \cref{fig:shasam_mine_res_minority} demonstrate imbalance with $A$ being the rare group, with distinguishable decision boundary between them.
Alongside selection of diverse anchors, \shasam-MINE selects hard-positives and negatives which continue to lie at the cluster boundary of $A$ and $B$ irrespective of the choice of target attribute label (anchors are mined from this set) - Rare class $A$ in \cref{fig:shasam_mine_res_minority} and abundant class $B$ in Fig. 4 of main paper. 
Minimizing $L_{\shasam}$ over the mined sets promotes reduction in  intra-group variance within the groups and maximizes inter-group separation between the target attribute and the remaining groups encouraging the model to learn features, unbiased to the sensitive attributes.

\noindent \textbf{Feature Similarity} (Cluster Overlap)
In contrast to distinct cluster boundaries depicted in \cref{fig:shasam_mine_res_balanced} and \cref{fig:shasam_mine_res_minority} we introduce a case with high inter-group bias demonstrated as overlapping feature clusters $A$ and $B$ in an imbalanced setting ($B$ being the abundant and $A$ being the rare group). The behavior of \shasam-MINE is consistent with the previous setting and mines hard positives and negatives at the cluster boundary. Interestingly, we see that \shasam-MINE selects hard-negatives that largely represent the overlapped section of the embedding space which when consumed by $L_{\shasam}$ would promote mitigation of inter-group bias and enforce strong decision boundaries between target attributes.

\subsection{Standard Deviation of Results on CelebA}
We indicate in Section 4.1 of the main paper that we report the average performance over three independent runs by varying the seed value of underlying libraries among three random seeds.
We supplement the results in Table 2 with standard deviation numbers in \cref{tab:results_celeba_with_std_dev} for \shasam\ variants and methods which were re-implemented by us. We show that models trained with \shasam\ achieves the lowest standard deviations among all compared approaches. This can be attributed to the combinatorial formulation which allows the model to learn from complete sets of anchors, positives and negatives rather than contrasting individual samples~\cite{tian2024fairvit} or one anchor pair and multiple positives and negatives~\cite{fscl}.

\begin{table*}[t]
\caption{\textbf{Classification results with standard deviations on CelebA} measured in terms of Top-1 Accuracy (Acc.) and equalized odds (EO) by varying the target $T$ and sensitive attributes $S$. Here, $a$, $b$, $e$, $m$, and $y$ denote attractiveness, big nose, bags-under-eyes, male, and young, respectively. All results are averaged over three independent runs. * indicates our re-implementations. All values are rounded off to 1 decimal point.}
\centering
\resizebox{\textwidth}{!}{
\begin{tabular}{l|cc|cc|cc|cc|cc|cc|cc|cc}
\toprule
\multirow{2}{*}{Method} & \multicolumn{2}{c|}{$T=a$ / $S=m$} & \multicolumn{2}{c|}{$T=a$ / $S=y$} & \multicolumn{2}{c|}{$T=b$ / $S=m$} & \multicolumn{2}{c|}{$T=b$ / $S=y$} & \multicolumn{2}{c|}{$T=e$ / $S=m$} & \multicolumn{2}{c|}{$T=e$ / $S=y$} & \multicolumn{2}{c|}{$T=e$ \& $b$ / $S=m$} & \multicolumn{2}{c}{$T=a$ / $S=m$ \& $y$} \\
       & EO ($\downarrow$) & Acc. ($\uparrow$) & EO ($\downarrow$) & Acc. ($\uparrow$) & EO ($\downarrow$) & Acc. ($\uparrow$) & EO ($\downarrow$) & Acc. ($\uparrow$) & EO ($\downarrow$) & Acc. ($\uparrow$) & EO ($\downarrow$) & Acc. ($\uparrow$) & EO ($\downarrow$) & Acc. ($\uparrow$) & EO ($\downarrow$) & Acc. ($\uparrow$) \\
\hline \hline
CE~\cite{ce_loss_nips}     & 27.8 & 79.6 & 16.8 & 79.8 & 17.6 & 84.0 & 14.7 & 84.5 & 15.0 & 83.9 & 12.7 & 83.8 & 12.9 & 72.6 & 31.3 & 79.5 \\
GRL~\cite{raff2018reversal}    & 24.9 & 77.2 & 14.7 & 74.6 & 14.0 & 82.5 & 10.0 & 83.3 & 6.7  & 81.9 & 5.9  & 82.3 & 9.4  & 71.4 & 22.9 & 78.6 \\
LNL~\cite{kim2019notlearn}    & 21.8 & 79.9 & 13.7 & 74.3 & 10.7 & 82.3 & 6.8  & 82.3 & 5.0  & 81.6 & 3.3  & 80.3 & 7.4  & 70.8 & 20.7 & 77.7 \\
FD-VAE~\cite{Park2021disentanglement} & 15.1 & 76.9 & 14.8 & 77.5 & 11.2 & 81.6 & 6.7  & 81.7 & 5.7  & 82.6 & 6.2  & 84.0 & 8.2  & 70.2 & 19.9 & 78.0 \\
MFD~\cite{jung2021distillation}    & 7.4  & 78.0 & 14.9 & 80.0 & 7.3  & 78.0 & 5.4  & 78.0 & 8.7  & 79.0 & 5.2  & 78.0 & 9.0  & 70.0 & 19.4 & 76.1 \\
SupCon~\cite{supcon} & 30.5 & 80.5 & 21.7 & 80.1 & 20.7 & 84.6 & 16.9 & 84.4 & 20.8 & 84.3 & 10.8 & 84.0 & 12.5 & 72.7 & 24.4 & 81.7 \\
\midrule
\multirow{2}{*}{FSCL (w/ group norm)~\cite{fscl}*} & 6.5  & 79.1 & 12.4 & 79.1 & 4.7  & 82.9 & 4.8  & 84.1 & 3.0  & 83.4 & \textbf{1.6}  & 83.5 & 2.5  & 70.8 & 17.0 & 77.2 \\
& \scriptsize$\pm$0.4  & \scriptsize$\pm$0.4 & \scriptsize$\pm$0.5 & \scriptsize$\pm$0.5 & \scriptsize$\pm$0.5 & \scriptsize$\pm$0.4 & \scriptsize$\pm$0.3  & \scriptsize$\pm$0.5 & \scriptsize$\pm$0.4  & \scriptsize$\pm$0.6 & \scriptsize$\pm$0.3  & \scriptsize$\pm$0.3 & \scriptsize$\pm$0.6  & \scriptsize$\pm$0.5 & \scriptsize$\pm$0.5 & \scriptsize$\pm$0.5 \\
% \rowcolor{LightGreen}
\multirow{2}{*}{\shasam (w/ LogDetCMI)}   & 6.1  & 80.7 & 10.5 & 78.6 & 3.6  & 84.5 & 4.2  & 85.9 & 2.7  & 85.0 & 1.7  & 84.0 & 2.5  & 71.3 & 15.8 & 78.7 \\
                        &\scriptsize$\pm$0.2  & \scriptsize$\pm$0.3 & \scriptsize$\pm$0.5 & \scriptsize$\pm$0.3 & \scriptsize$\pm$0.6 & \scriptsize$\pm$0.2 & \scriptsize$\pm$0.3  & \scriptsize$\pm$0.3 & \scriptsize$\pm$0.2  & \scriptsize$\pm$0.2 & \scriptsize$\pm$0.6  & \scriptsize$\pm$0.4 & \scriptsize$\pm$0.2  & \scriptsize$\pm$0.6 & \scriptsize$\pm$0.4 & \scriptsize$\pm$0.3 \\
% \rowcolor{LightGreen}
\multirow{2}{*}{\shasam (w/ FLCMI)}   & \textbf{5.5}  & \textbf{81.3} & \textbf{9.9}  & \textbf{79.6} & \textbf{3.3}  & \textbf{84.7} & \textbf{3.9}  & \textbf{87.0} & \textbf{2.6}  & \textbf{85.8} & \textbf{1.6}  & \textbf{84.4} & \textbf{2.2}  & \textbf{71.8} & \textbf{14.6} & \textbf{79.5} \\
                   & \scriptsize$\pm$0.3  & \scriptsize$\pm$0.4 & \scriptsize$\pm$0.7 & \scriptsize$\pm$0.4 & \scriptsize$\pm$0.6 & \scriptsize$\pm$0.4 & \scriptsize$\pm$0.5  & \scriptsize$\pm$0.3 & \scriptsize$\pm$0.3  & \scriptsize$\pm$0.7 & \scriptsize$\pm$0.4  & \scriptsize$\pm$0.2 & \scriptsize$\pm$0.4  & \scriptsize$\pm$0.5 & \scriptsize$\pm$0.3 & \scriptsize$\pm$0.5 \\
\midrule
\multirow{2}{*}{FSCL (w/ ViT)~\cite{fscl}}        & 5.7  & 79.9 & 10.8 & 80.0 & 4.0  & 85.0 & 4.6  & 88.2 & 2.9  & 87.7 & 1.7  & 86.3 & 2.5  & 72.8 & 16.3 & 81.8 \\
                        & \scriptsize$\pm$0.4  & \scriptsize$\pm$0.2 & \scriptsize$\pm$0.5 & \scriptsize$\pm$0.3 & \scriptsize$\pm$0.3 & \scriptsize$\pm$0.3 & \scriptsize$\pm$0.4  & \scriptsize$\pm$0.6 & \scriptsize$\pm$0.4  & \scriptsize$\pm$0.5 & \scriptsize$\pm$0.5  & \scriptsize$\pm$0.5 & \scriptsize$\pm$1.0  & \scriptsize$\pm$1.1 & \scriptsize$\pm$0.6 & \scriptsize$\pm$0.6 \\
\multirow{2}{*}{FairViT~\cite{tian2024fairvit}*}  & 6.4  & 83.8 & 12.6 & 82.5 & 5.1  & 86.1 & 4.7  & 89.3 & 3.4  & 88.1 & 1.8  & \textbf{86.8} & 2.9  & \textbf{74.4} & 17.4 & 82.4 \\
                           & \scriptsize$\pm$0.2  & \scriptsize$\pm$0.4 & \scriptsize$\pm$0.2 & \scriptsize$\pm$0.1 & \scriptsize$\pm$0.4 & \scriptsize$\pm$0.1 & \scriptsize$\pm$0.2  & \scriptsize$\pm$0.1 & \scriptsize$\pm$0.3  & \scriptsize$\pm$0.3 & \scriptsize$\pm$0.3  & \scriptsize$\pm$0.3 & \scriptsize$\pm$0.8  & \scriptsize$\pm$0.9 & \scriptsize$\pm$0.5 & \scriptsize$\pm$0.5 \\
% \rowcolor{LightGreen}
\multirow{2}{*}{\shasam (w/ FLCMI, ViT)} & \textbf{5.3}  & \textbf{84.0} & \textbf{9.9}  & \textbf{82.3} & \textbf{3.0}  & \textbf{85.6} & \textbf{3.7}  & \textbf{89.2} & \textbf{2.8}  & \textbf{88.2} & \textbf{1.6}  & 86.8 & \textbf{2.2}  & 74.3 & 14.6 & \textbf{82.4} \\
                    & \scriptsize$\pm$0.2  & \scriptsize$\pm$0.3 & \scriptsize$\pm$0.4 & \scriptsize$\pm$0.1 & \scriptsize$\pm$0.3 & \scriptsize$\pm$0.2 & \scriptsize$\pm$0.2  & \scriptsize$\pm$0.2 & \scriptsize$\pm$0.3  & \scriptsize$\pm$0.3 & \scriptsize$\pm$0.2  & \scriptsize$\pm$0.3 & \scriptsize$\pm$0.7  & \scriptsize$\pm$1.0 & \scriptsize$\pm$0.6 & \scriptsize$\pm$0.4 \\
\bottomrule
\end{tabular}}
\label{tab:results_celeba_with_std_dev}
\end{table*}

\section{Limitations}
\label{sec:limitations}
While the \shasam\ framework presents a novel combinatorial approach for fair facial attribute recognition, it has a few limitations. The primary constraint is the computational cost and wall-clock time overhead associated with the \shasam-MINE stage, which involves combining submodular optimization and representation learning in a unified framework. To remedy this, \shasam\ relies on training with a randomly selected subset of the dataset at each epoch, which could be inefficient for extremely large-scale applications.
Further, \shasam\ operates on the assumption that discrete target and sensitive attributes are available, and in experiments, continuous attributes like 'age' were simplified into binary categories, a step that may not be suitable for all real-world scenarios.
Finally, complex submodular functions like Log-Determinant may require computing large matrices (given a large batch size) which might expand the compute requirements beyond the available budget.

\end{document}